%% file: main.tex
\documentclass[10pt,twocolumn,letterpaper]{article}

\usepackage[pagenumbers]{cvpr} % To force page numbers, e.g. for an arXiv version

\input{preamble}

\definecolor{cvprblue}{rgb}{0.21,0.49,0.74}
\usepackage[pagebackref,breaklinks,colorlinks,citecolor=cvprblue]{hyperref}
\usepackage{multirow}
\usepackage{makecell}
\usepackage{xcolor,colortbl}
\usepackage{color}
\usepackage{comment}
\usepackage[misc]{ifsym}

\definecolor{minetable1colorx}{rgb}{0.75, 0.75, 0.75}

\definecolor{lightgrey}{gray}{0.9}

\usepackage{algorithm}
\usepackage{algorithmic}

\title{Improving Batch Normalization with TTA for \\Robust Object Detection in Self-Driving}

\author{%
\textbf{Dacheng Liao, Mengshi Qi, Liang Liu, Huadong Ma} \\ 
State Key Laboratory of Networking and Switching Technology~~~\\ Beijing University of Posts and Telecommunications, China\\
\{liaodacheng, qms, liangliu, mhd\}@bupt.edu.cn
}

\begin{document}
\maketitle

\begin{abstract}
In current open real-world autonomous driving scenarios, challenges such as sensor failure and extreme weather conditions hinder the generalization of most autonomous driving perception models to these unseen domain due to the domain shifts between the test and training data. As the parameter scale of autonomous driving perception models grows, traditional test-time adaptation (TTA) methods become unstable and often degrade model performance in most scenarios. To address these challenges, this paper proposes two new robust methods to improve the Batch Normalization with TTA for object detection in autonomous driving: (1) We introduce a LearnableBN layer based on Generalized-search Entropy Minimization (GSEM) method. Specifically, we modify the traditional BN layer by incorporating auxiliary learnable parameters, which enables the BN layer to dynamically update the statistics according to the different input data. (2) We propose a new semantic-consistency based dual-stage-adaptation strategy, which encourages the model to iteratively search for the optimal solution and eliminates unstable samples during the adaptation process. Extensive experiments on the NuScenes-C dataset shows that our method achieves a maximum improvement of about 8\% using BEVFormer as the baseline model across six corruption types and three levels of severity. We will make our source code available soon.
\end{abstract}
\section{Introduction}
Autonomous driving perception models encounter significant challenges when the distribution of test data diverges from that of the training data, particularly in dynamic and open real-world driving scenarios~\cite{bojarski2016end} such as extreme weather conditions or sensor failures, leading to severe degradation in the model's predictive accuracy~\cite{zhu2023understanding}, which is unacceptable for autonomous driving tasks. Traditional methods~\cite{yun2019cutmix,devries2017improved,zhang2017mixup} for enhancing model robustness typically rely on extensive annotation costs or use data augmentation. However, these methods necessitate prior knowledge of the test data distribution, which is often unknown in real-world driving scenarios.
To address these practical issues, a more viable approach is to use TTA methods~\cite{zhang2023adanpc} to adjust models promptly when facing unseen domains. 

The prevalent TTA paradigm~\cite{boudiaf2022parameter} typically addresses the issue of the distribution shifts between test and training data by adjusting the statistics of the Batch Normalization (BN) layers, As shown in Fig~\ref{figure1}.
However, this TTA paradigm presenting the following challenges in self-driving~\cite{liang2024comprehensive}:

\begin{figure}
    \centering
    \includegraphics[width=1\linewidth]{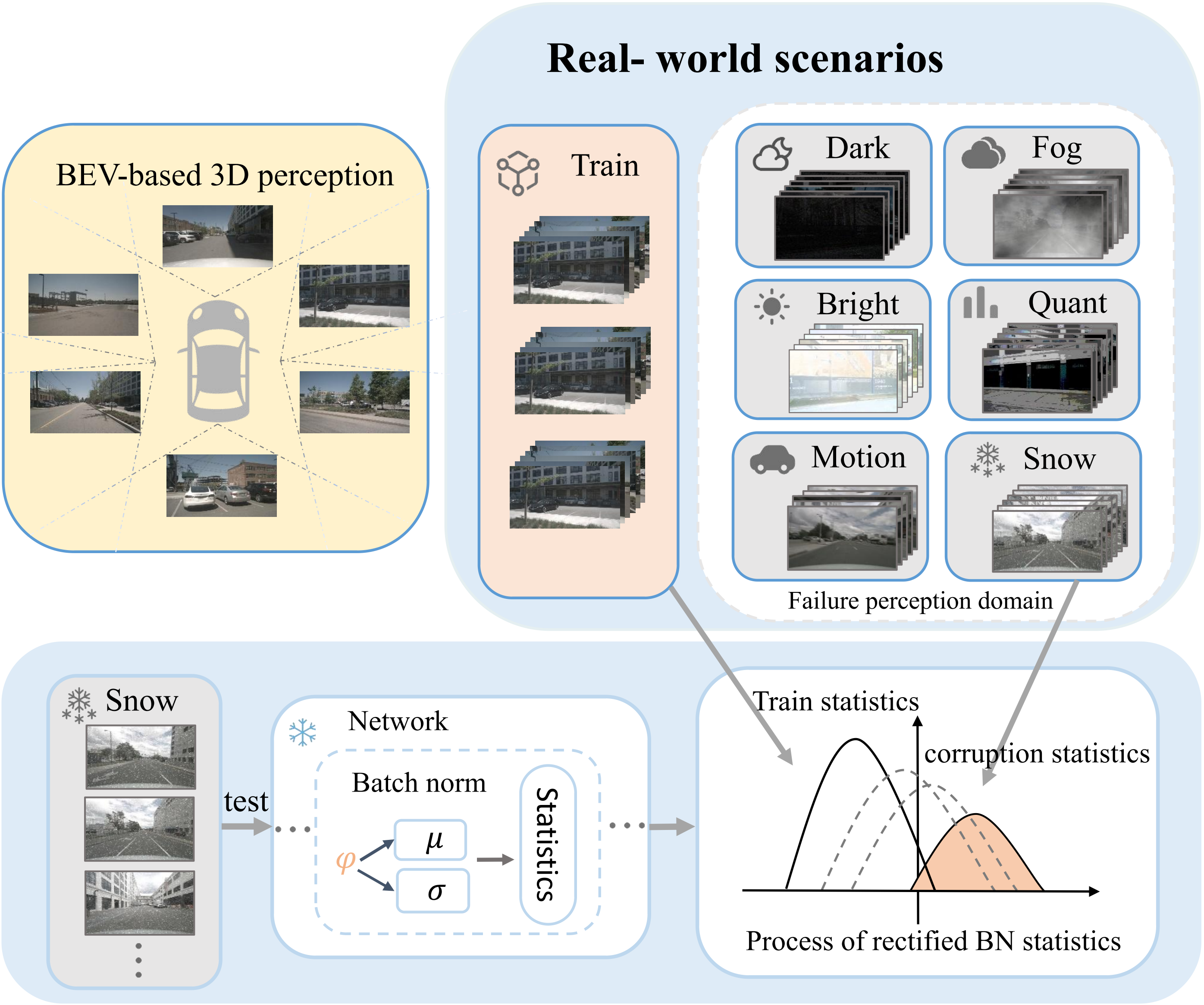}
    \caption{Illustration of the problems faced by BEV-based 3D object detection model struggles to perceive unseen domains caused by extreme weather conditions. In order to enhance the robustness of the model, TTA method estimating the BN statistics of the unseen domains during the testing phase.}
    \label{figure1}
\end{figure}

Firstly, TTA methods that adjust Batch Normalization (BN) parameters exhibit significant instability in autonomous driving perception tasks due to the BN layers employ an exponential moving average (EMA) approach to estimate the data distribution. The EMA method is highly sensitive to batch size, meanwhile the use of EMA for updating BN statistics is significantly affected by the problem of internal covariate shift in the model. If the prediction of the bottom BN layer's statistic is error, it can lead to the accumulation of errors in subsequent BN layers' prediction~\cite{schneider2020improving}. As model parameters and depth increase in autonomous driving perception tasks, the batch size is constrained, making it difficult for TTA methods to accurately predict the real test data distribution and worsening internal covariate shift.

Furthermore, TTA methods~\cite{benz2021revisiting} that employ unsupervised method, such as entropy minimization (EM)~\cite{wang2020tent}, are also commonly used. These methods presents a potential issue of error accumulation. During model optimization, the absence of ground truth annotations often causes the direction of the gradient in the parameter space to be influenced by the direction of historical gradients, leading to increased model confidence that deviates from the true solution. If the model’s initial state is not ideal, entropy minimization may lead the model to optimize towards a degeneration to a trivial solution.

Additionally, TTA methods~\cite{li2023robustness} typically classify test samples first and then use samples within specific categories to adjust the model. This requires prior knowledge of the distribution types within the test data. In real-world driving scenarios, the diversity of encountered scenes is often unknown, and the presence of noisy samples is prevalent~\cite{niu2023towards}.

To address these challenges, we propose to improve the Batch Normalization with TTA for robust object detection in self-driving. Firstly, we introduce a learnable batch normalization layer and generalized search entropy minimization to adjust BN statistics. By introducing auxiliary learnable parameters into BN layers, we can predict the BN statistics of the test domain using these parameters, replacing the EMA method. This approach addresses the limitations of BN layers under mini-batch conditions, mitigates model internal covariate shift issues and addresses the instability arising from adjusting BN statics. Additionally, by guiding the optimizing of auxiliary learnable parameters through entropy minimization, we introduce the generalized searches to mitigating the limitations of entropy minimization. Secondly, to tackle the challenges of TTA in real-world scenarios, we propose a semantic-consistency based dual-stage-adaptation method. By adjusting the variation of learning rates and dividing adaptation into two stages, we use the semantic consistency of sample predictions in different stages as guidance to filter out the uncertain samples, thereby making the training process more stable and prevent the model from converging to a local optimum in the solution space.

Our main contributions can be summarized as follows:
\begin{enumerate}
\item %For the first time, We apply the TTA method based on adjusted BN statistics to bev-based perception for automated driving. 
We propose a novel TTA paradigm for robust BEV perception in open real-world driving scenarios, by incorporating a LearnableBN for estimating BN statistics and generalized search entropy minimization (GSEM) loss function that effectively addresses the instability issues inherent to traditional BN layers. 
\item We introduce a semantic-consistency based dual-stage-adaptation method, which is designed to filters out the noisy samples and prevents the model from converging to a local optimum in the solution space.
% \item We propose LearnableBN method to estimate BN statistics. By introducing auxiliary learnable parameters effectively addresses the instability issues inherent to traditional BN layers.
% \item We propose generalized search entropy minimization(GSEM) loss function, which designed to address the uncertainty issues that arise in the absence of ground truth annotations in unsupervised learning.
\item We conduct extensive experiments on widely-adopted benchmark, nuScenes-C, and results show that our proposed method achieves a maximum improvement of about 8\% using BEVFormer as the baseline model across six corruption types and three levels of severity.

\end{enumerate}
\section{Related Work}

\textbf{Autonomous driving perception} primarily focuses on 3D object detection. In monocular 3D object detection~\cite{zou2021devil}, some methods use additional pre-trained depth estimation modules to address one of the most challenging problems in Mono 3Det~\cite{ding2020learning}, which is depth estimation from a single image. SMOKE~\cite{liu2021ttt++} proposes treating 3D object detection as a keypoint estimation task. Later, Monoflex~\cite{zhang2021objects} improves this approach by providing a flexible definition of object centers, unifying the centers of regular and truncated objects. GrooMeD-NMS~\cite{kumar2021groomed} introduces a grouped mathematically differentiable Non-Maximum Suppression method for Mono 3Det.

The mainstream approach for BEV(bird eye view) based object detection involves Object query-based algorithms, including: DETR3D~\cite{wang2022detr3d}, which leverages Transformer’s cross-attention mechanism to avoid explicit depth estimation. PETR, which enhances performance by constructing 3D position-aware representations. BEVFormer~\cite{li2022bevformer,yang2023bevformer}, which employs temporal cross-attention and uses polar coordinates for object detection. Sparse4D~\cite{lin2022sparse4d}, which uses sparse proposals for feature fusion. To validate the generality of the method. In this paper, we select BEVFormer, Sparse4D, MOnofelx as our baseline models to test the effectiveness of TTA methods in real-world scenarios.

\noindent\textbf{Test-Time-Adaptation (TTA)}~\cite{sun2020test,fleuret2021test,iwasawa2021test} aim to fine-tune models on unlabeled test images during the testing phase. In the work by Benz et al. \cite{benz2021revisiting} proposes a method that adjusts BN statistics during testing through forward propagation without additional training. Schneider et al~\cite{schneider2020improving} propose dynamically calculating the mixture coefficient based on the quantities used to predict the test BN statistics. TENT~\cite{wang2020tent} is an unsupervised learning method that first proposed using entropy minimization as singular loss function to estimate BN statistics and optimizes channel-wise affine transformations. Following the TENT method, Domain adaptor~\cite{zhang2023domainadaptor} dynamically computes mixture coefficient in EMA method and uses temperature scaling to optimize entropy minimization loss.

\begin{figure*}
    \centering
    \includegraphics[width=0.95\linewidth]{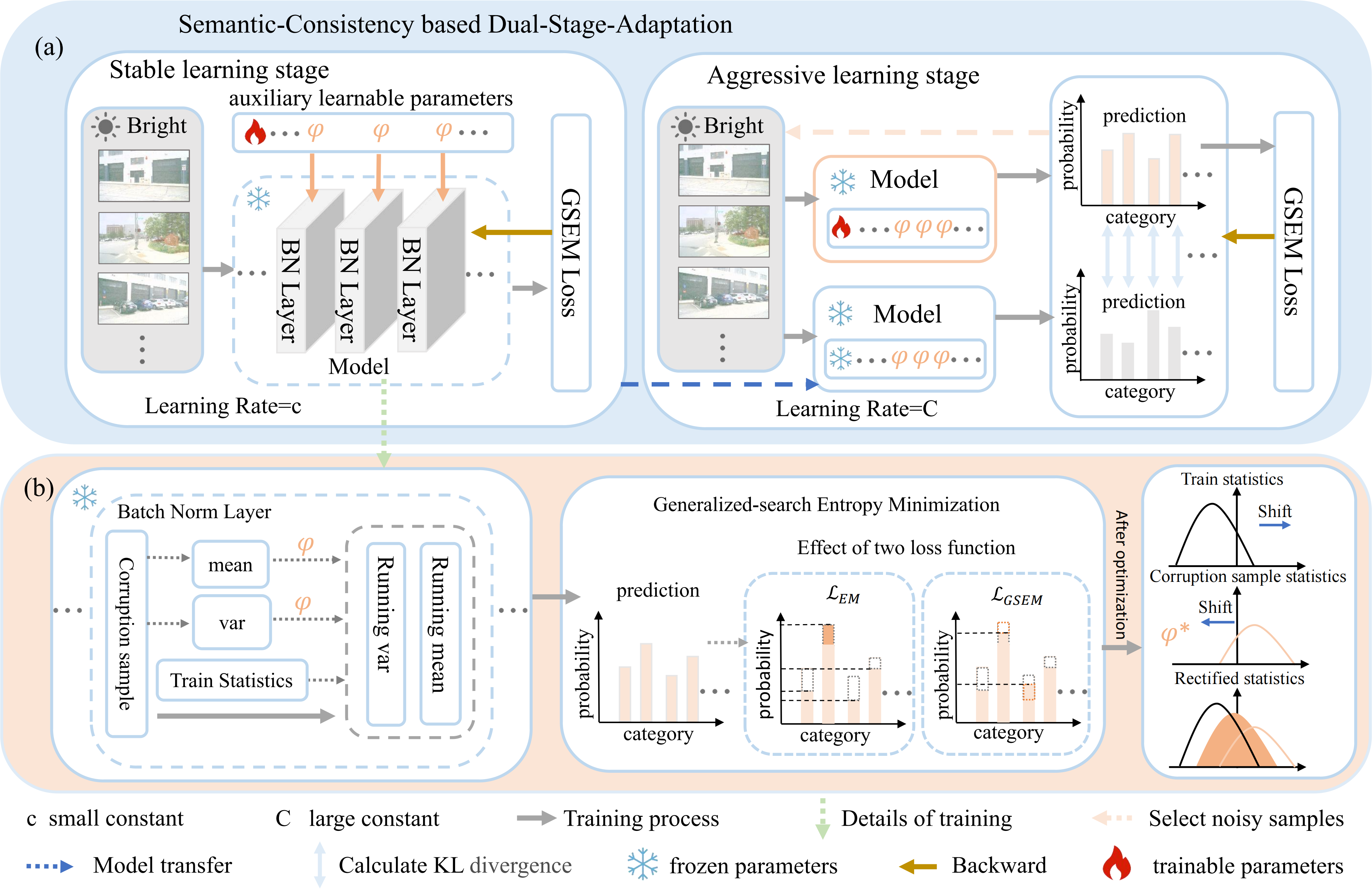}
    \caption{\textbf{Method Overview}. Module (a) demonstrates the Semantic-Consistency based Dual-Stage-Adaptation, which consists of a stable adaptation phase with a low learning rate and an aggressive adaptation phase with a high learning rate. In the aggressive adaptation phase, the model trained in the stable adaptation phase is used to predict the same samples, and calculate KL divergence between their prediction to filter noisy samples. Module (b) is intended to describe the training process. First, auxiliary learnable parameters are introduced into the BN layer, We frozen all model parameters, and only the auxiliary parameters are learnable. Then adaptation is conducted using the GSEM loss function. It is important to note that the BN statistics are not changed during forward propagation, but are rectified after optimization. }
    \label{figure2}
\end{figure*}

\section{Method}
\subsection{Problem Definition}
In this work, we define the test dataset as $D_c^s=\{x_1,x_2,...,x_n\}$, where $c$ represents the different conditions in real-world driving scenarios, and $s$ is the severity level of the domain shift between test domain and train domain. We define the model as $f(\cdot|\theta,\phi)$, where the $\theta$ is origin model's parameters. We introduce the set of auxiliary learnable parameters in the BN layers, defined as $\phi=\{\phi_1,\phi_2,...,\phi_m\}$ where $m$ corresponds to the parameters for the $m$-th BN layer. The learning rate of the model is defined as $\eta$.

\subsection{Overview}
Our TTA method apply two stage adaptation to predict the BN statistics $(\mu, \sigma)$ of test domain $D_c^s$ in each BN layer. Specifically, we use generalized-search entropy minimization as the loss function $\mathcal{L}_{GSEM}$ to optimize the learnale mixture coefficient $\phi$ that we introduced in the BN layer. After each step of optimizing, we perform secondary correction on the BN statistics $(\mu, \sigma)$ using the optimized $\phi$. Additionally, we propose a semantic-consistency based dual-stage-adaptation method. The first stage is the stable adaptation stage, which employs a smaller learning rate $\eta$ with the aim of conservatively estimating the BN statistics. The second stage is the aggressive adaptation stage, using a larger learning rate $\eta$ to help the $\phi$ escape local optima and converge to global optima. To ensure the stability of model adaptation, the predictions from the second stage are compared semantic consistency with the predictions from the first stage. This comparison is used to filter noisy samples from the test domain $D_c^s$. The whole framework is illustrated in Fig.~\ref{figure2}.

\subsection{LearnableBN}
\subsubsection{Generalized-search Entropy Minimization.}~Tent~\cite{wang2020tent} proposes an approach to employ entropy minimization as the singular loss function for test-time-adaptation. Relying solely on entropy minimization loss presents several challenges. During training, the gradients of the entropy minimization tend to amplify as the loss decreases, thereby rendering the model is susceptible to collapsing into trivial solutions. 
Furthermore, in the absence of annotated data in unsupervised training, it becomes difficult to ascertain whether the label with the highest confidence is indeed correct, potentially leading the model to become overly confident in incorrect predictions.

Prevailing strategies address the limitation of entropy minimization by introducing a temperature coefficient to reduce the sharp distribution. These methods does not alter the model's original semantic information. In real-world scenarios, there is a significant likelihood that the model's original semantic information may be erroneous.
Consequently we introduce the generalized-search entropy minimization loss:
\begin{equation}
\mathcal{L}_{GSEM}=\mathcal{L}_{EM}+\mathcal{L}_{GS}
\label{eqGSEM}
\end{equation}
Generalized-search entropy minimization loss consists of two parts: the first part is entropy minimization loss $\mathcal{L}_{EM}$ , and the second part is a regularization loss $\mathcal{L}_{GS}$ used to modify the gradient direction of the entropy minimization loss. The formulas for the two losses are as follows:
\begin{align}
\mathcal{L}_{EM}&=\sum_Q^i\sum_C^j{-p_{i,j}\log{p_{i,j}}}\\
\mathcal{L}_{GS}&=\sum_Q^i\max \limits_{j\in C}p_{i,j}-\min \limits_{j\in C}p_{i,j}
\label{eqLGS}
\end{align}
where $Q$ is the query numbers, $C$ is the numbers of classes, $p_{i,j}$ is the predicted probability of the different classes of query.

We propose a regularization loss $\mathcal{L}_{GS}$. As shown in Eq.~\ref{eqLGS}, $\mathcal{L}_{GS}$ is designed to penalize the divergence between the model's highest probability prediction and its lowest probability prediction for a given query. It aims to mitigate the issue of increasing gradient magnitude as the loss decreases during entropy minimization and balancing the contribution of model's different class predictions to the loss. This helps prevent the model from converging to trivial solutions. 

Additionally, to mitigate the impact of uncertainty in model predictions on model adapting, $\mathcal{L}_{GS}$ introduces perturbations to the model's gradients, allowing optimization process without entirely relying on maximum gradients direction and reduce the impact of historical gradient directions on the current gradient direction of the model. This helps model in escaping local optima during training and explore a broader solution space, thereby enhancing model's generalization ability. 

Simultaneously, Using entropy minimization loss to directly adjust model parameters can amplify the impact of erroneous predictions on model adapting. In LearnableBN method, $\mathcal{L}_{GSEM}$ loss is used to optimize the auxiliary learnable parameters, denoted as $f\phi$, which we introduce. These parameters can indirectly predict the BN statistics.

\subsubsection{Optimizing BN layers.}~The inherent instability of the BN layer is mainly attributed to the following factors: 

(1) The exponential moving average (EMA) method used to predict statistics in the BN layer is highly dependent on batch size. If the batch size is too small, it might not accurately reflect the full distribution of the test data domain, potentially leading to erroneous shifts in BN statistics.

(2) Within the neural network, deeper layer information is found to exhibit greater transferability, while shallow layers information often requires more frequent updates. Therefore, the update strategy for BN statistics should be different for each layer.

(3) Predictions of BN statistics are highly sensitive to internal covariate shift, where the accuracy of statistical predictions in deep BN layers significantly influences those in shallow BN layers.

Therefore, we propose a novel BN layer method for predicting BN statistics to replace the EMA method:

In $m$-th BN layer the equation in forward propagation:
\begin{equation}
\mu=(1-\phi_m)\mu_{h}+\phi\mu_{p}
\label{eqforwardmu}
\end{equation}
\begin{equation}
\sigma^2=(1-\phi_m)(\sigma_{h})^2+\phi(\sigma_{p})^2
\label{eqforwardsigma}
\end{equation}
\begin{equation}
\hat{z}=\frac{z-\mu}{\sqrt{\sigma^2+\epsilon}}\gamma+\beta
\end{equation}

\noindent where $z$ and $\hat{z}$ represent the input and outputs of BN layer. ($\mu_{h},\sigma_{h})$ represent the history BN statistics and $(\mu_{p},\sigma_{p})$ represent the BN statistics calculated from present sample. $(\gamma,\beta)$ is the affine parameters of the BN layer. $\epsilon$ is a small constant added to ensure numerical stability. We introduce a new learnable parameter $\phi_m$ to each BN layer and apply the leakyrelu function with a hyperparameter of -0.001 after $\phi_m$ to avoid negative values. This enables each BN layers to have independent mixture coefficient. At this stage, the BN statistic $(\mu_{p},\sigma_{p})$ calculated from Eq.~\ref{eqforwardmu} and Eq.~\ref{eqforwardsigma} are utilized as a temporary variables to influence the model's predictions. 

After optimizing with $\mathcal{L}_{GSEM}$, we introduce a quadratic correction:
\begin{equation}
\phi_m^{(t+1)}=\phi_m^{(t)}-\eta\cdot\nabla_{\phi}\mathcal{L}_{GSEM}
\label{optimise}
\end{equation}
\begin{equation}
\mu=(1-\phi_m^{(t+1)})\mu_{h}+\phi_m^{(t+1)}\mu_{p}
\label{secondrevisemu}
\end{equation}
\begin{equation}
\sigma^2=(1-\phi_m^{(t+1)})(\sigma_{h})^2+\phi_m^{(t+1)}(\sigma_{p})^2
\label{secondrevisesigma}
\end{equation}
Where the $\{...,t-1,t,t+1,...\}$ represent each optimization step in the the training iterations. 
The first correction is necessary because the BN layer dynamically mixes the current sample's statistics with history statistics during the prediction process, helping to reduce domain shift and enabling the model to predict the mixture coefficient for current sample more accurately.  

The second revision is due to the delay in the impact of the $\phi_m$ on the BN statistics. The $\phi_m$ after optimisation should be the mixing coefficients of the current samples. If we use Eq.~\ref{eqforwardmu} and Eq.~\ref{eqforwardsigma} as the BN statistic, it will result in the $\phi_m$ optimized by current $\mathcal{L}_{GSEM}$ to be used in the next sample's mixing coefficients. Therefore, we made specific adjustments to the model training proces. After optimizing $\phi_m$ using Eq.~\ref{optimise}, we applied Eq.~\ref{secondrevisemu} and Eq.~\ref{secondrevisesigma} to correct the statistics of the BN layers.

We propose a method to optimize the BN layer by introducing a auxiliary learnable parameters $\phi_m$ to replace the EMA method. It mitigates the limitation where the accuracy of BN statistics predicted using the EMA method is highly dependent on batch size, resulting in a more stable process for predicting BN statistics. Applying different BN statistics shift strategies for each BN layers, effectively utilized the transferability of the deep BN layers.
It is worth noting that, unlike the traditional model parameter, $\phi_m$ is an auxiliary parameter that will initialized at the start of each domain adaptation, enabling specific adaptation strategies for different domains.

\subsection{Semantic-Consistency based Dual-Stage-Adaptation}
In our LearnableBN method only the auxiliary learnable parameters $\phi$ are optimized. Adapting with a very small learning rate often results in the model converging to a local optimum due to the limited number of trainable parameters. Conversely, the peculiarities of $\mathcal{L}_{GSEM}$ can cause the model to converge to a trivial solution if an excessively large learning rate is used. To further enhance the generalization of our method and effectively handle noisy samples encountered in real-world scenarios. We propose a semantic-consistency based dual-stage-adaptation method.

First of all, in the first stage, a small learning rate is used to allow the model to find the local optimum. In the second stage, we use a large learning rate to allow the model to escape from the local optimum. 
In order to guarantee the reliability of the adapting process, we compare the semantic consistency between the first-stage model and the second-stage model by using both models to predict the same sample and then comparing the Kullback-Leibler divergence (KL) between their predictions. We consider a sample to be stable for model adapting if the KL value is in the lowest 10\% of historical KL values. More algorithm details are put in Appendix.

The rationale behind the first adapting stage is that the model often exhibits instability when confronted with unseen domains. Therefore, the original model cannot be used directly as a semantic comparison model. The local optimums obtained by the model in the first adapting stage are more transferable. Consequently, we use the predictive power of the local optimums to filter the unstable samples. During the second stage of adapting, the learning rate is increased  in order to encourage the model to converge to the global optimum. 

Concurrently, a semantic consistency based method is used for sample selection, which considers the hidden layer features of samples. This approach guarantees the  adapting stability while minimizing the risk of learning noisy samples during test-time-adaptation.

\begin{table*}[htbp] 
\centering
\small
\setlength{\tabcolsep}{6pt}
\begin{tabular}{l|c|c|cc|c|c|cc|c|c|cc}
\toprule
\toprule
\multicolumn{1}{l|}{\textbf{}} & \multicolumn{4}{c|}{\textbf{Low Light}} & \multicolumn{4}{c|}{\textbf{Fog}} & \multicolumn{4}{c}{\textbf{Motion blur}} \\ 
\midrule
\multicolumn{1}{l|}{\textbf{Severity}} & \multicolumn{1}{c}{\textbf{Easy}} & \multicolumn{1}{c}{\textbf{Mid}} & \multicolumn{1}{c}{\textbf{Hard}} & \textbf{Avg} & \multicolumn{1}{c}{\textbf{Easy}} & \multicolumn{1}{c}{\textbf{Mid}} & \multicolumn{1}{c}{\textbf{Hard}} & \textbf{Avg} & \multicolumn{1}{c}{\textbf{Easy}} & \multicolumn{1}{c}{\textbf{Mid}} & \multicolumn{1}{c}{\textbf{Hard}} & \textbf{Avg} \\ 
%\midrule
%\multicolumn{13}{|c|}{This is a merged row across all columns} \\ 
\midrule
 
\multicolumn{1}{l|}{\textit{Baseline}} & \multicolumn{1}{c}{\underline{0.4011}} & \multicolumn{1}{c}{\underline{0.3352}} & \underline{0.2274} & \underline{0.3212} & \multicolumn{1}{c}{\textbf{0.4908 }} & \multicolumn{1}{c}{\underline{0.4825}} & \underline{0.4655} & \underline{0.4796} & \multicolumn{1}{c}{\underline{0.4661}} & \multicolumn{1}{c}{0.3002} & 0.2328 & 0.3330\\ 
 
\multicolumn{1}{l|}{ReviseBN} & \multicolumn{1}{c}{0.3382} & \multicolumn{1}{c}{0.2798} & 0.1715 & 0.2631 & \multicolumn{1}{c}{0.4296} & \multicolumn{1}{c}{0.4188} & 0.4048 & 0.4177 & \multicolumn{1}{c}{0.4316} & \multicolumn{1}{c}{\underline{0.3444 }} & \underline{0.2905 } & \underline{0.3555} \\ 
 
\multicolumn{1}{l|}{TENT} & \multicolumn{1}{c}{0.2636} & \multicolumn{1}{c}{0.2085} & 0.149 & 0.2070 & \multicolumn{1}{c}{0.3416} & \multicolumn{1}{c}{0.333} & 0.3161 & 0.3323 & \multicolumn{1}{c}{0.3185} & \multicolumn{1}{c}{0.1842} & 0.1442 & 0.2823 \\

\multicolumn{1}{l|}{AdaBn} & \multicolumn{1}{c}{0.104} & \multicolumn{1}{c}{0.075} & 0.0528 & 0.0772 & \multicolumn{1}{c}{0.1388} & \multicolumn{1}{c}{0.1345} & 0.1266 & 0.1333 & \multicolumn{1}{c}{0.1325} & \multicolumn{1}{c}{0.1296} & 0.1218 & 0.1279  \\

\multicolumn{1}{l|}{ARM(BN)} & \multicolumn{1}{c}{0.1319} & \multicolumn{1}{c}{0.0978} & 0.0587 & 0.0961 & \multicolumn{1}{c}{0.1473} & \multicolumn{1}{c}{0.1449} & 0.1372 & 0.1431 & \multicolumn{1}{c}{0.1621} & \multicolumn{1}{c}{0.1535} & 0.1297 & 0.1484\\ 
 
\midrule
\multicolumn{1}{l|}{LearnableBN} & \multicolumn{1}{c}{\textbf{0.4069}} & \multicolumn{1}{c}{\textbf{0.3585}} & \textbf{0.2753} & \textbf{0.3469} & \multicolumn{1}{c}{\underline{0.4899}} & \multicolumn{1}{c}{\textbf{0.4829 }} & \textbf{0.4697 }& \textbf{0.4808} & \multicolumn{1}{c}{\textbf{0.4698}} & \multicolumn{1}{c}{\textbf{0.3567}} & \textbf{0.3098} & \textbf{0.3787} \\  
\bottomrule
\bottomrule
\end{tabular}
\caption{Comparison of different TTA methods on \textbf{Nuscenes-C} across three levels of severity. The baseline model is \textbf{BEVFormer} with \textbf{ResNet-101} as the backbone. \textbf{Bold}: Best in the category. \underline{Underline}: Second best in the category.}
\label{baselinetable2}
\end{table*}

\section{Experiments}
\subsection{Experimental Setup}

\textbf{Dataset.} To simulate a dynamic and real-world autonomous driving scenarios, the experiments are conducted on the Nuscenes-C dataset and Kitti-C dataset. NuScenes-C adds natural corruption, including exterior environments, interior sensors factors, and temporal factors, based on NuScenes~\cite{caesar2020nuscenes}. It includes six types of corruption and three levels of severity: EASY, MID, and HARD.
The KITTI-C dataset introducing 12 distinct types of data corruptions to the validation set based on KITTI dataset~\cite{geiger2013vision}.
Our method compares with TTA methods, without introducing additional source data and without relying on annotations.

\noindent\textbf{Metrics.} In the BEV based 3D object detection task, We evaluate the performance of our method with the official nuScenes mertric, \emph{nuScenes Detection Score (NDS)}, which calculating a weighted sum of mAP, mATE, mASE, mAOE, mAVE, and mAAE. For the monocular 3D object detection task, we present our experimental results in terms of Average Precision (AP) for 3D bounding boxes, denoted as $AP_{3D|R_{40}}$. More details please refer to our supplementary materials.

\noindent\textbf{Implementation Details.} We implement our model based on Pytorch on a single NVIDIA L20 GPU. The baseline models used are BEVFormer~\cite{li2022bevformer}, Sparse4D~\cite{lin2022sparse4d} and MonoFlex.
In Nuscenes-C, to evaluate the stability of TTA methods, the batch size was set to 1. In the semantic-consistency-based dual-stage-adaptation, we set the learning rates $\eta$ to 2e-8 and 2e-7 and learning ratio $\alpha$ set at 0.1. The initial value of auxiliary learnable parameters $\phi$ in BN layers is set to 1e-5. The implementation details of Kitti-C are put in supplementary materials.

\begin{table*}[htbp] 
\centering
\small
\setlength{\tabcolsep}{6pt}
\begin{tabular}{l|c|c|c|cc|c|ccc|ccc}
\toprule
\toprule
\multicolumn{1}{l|}{\textbf{}} & \multicolumn{3}{c|}{\textbf{Noise}} & \multicolumn{3}{c|}{\textbf{Blur}} & \multicolumn{3}{c|}{\textbf{Weather}}& \multicolumn{3}{c}{\textbf{Digital}} \\ 
\midrule
\multicolumn{1}{l|}{\textbf{Method}} & \multicolumn{1}{c}{\textbf{Gauss.}} & \multicolumn{1}{c}{\textbf{Shot}} &\multicolumn{1}{c|}{ \textbf{Impul.}} & \multicolumn{1}{c}{\textbf{Defoc.}} & \multicolumn{1}{c}{\textbf{Glass}} & \multicolumn{1}{c|}{\textbf{Motion}} & \multicolumn{1}{c}{\textbf{Frost}} & \multicolumn{1}{c}{\textbf{Fog}} & \multicolumn{1}{c|}{\textbf{Brit.}} & \multicolumn{1}{c}{\textbf{Contr.}} & \multicolumn{1}{c}{\textbf{Pixel.}} & \multicolumn{1}{c}{\textbf{Sat.}} \\ 
%\midrule
%\multicolumn{13}{|c|}{This is a merged row across all columns} \\ 
\midrule
 
\multicolumn{1}{l|}{\textit{Baseline}} & \multicolumn{1}{c}{0.19} & \multicolumn{1}{c}{1.62} & 0.32 & 3.72 & \multicolumn{1}{c}{8.47 } & 6.22 & 4.27 & \multicolumn{1}{c}{2.25} & 9.19 & \multicolumn{1}{c}{2.08} & 1.83&9.11 \\ 
 
\multicolumn{1}{l|}{BN adaptation} & \multicolumn{1}{c}{6.21} & \multicolumn{1}{c}{8.20} & 9.20 & 7.83 & \multicolumn{1}{c}{5.35} & 7.52 & 6.47 & 9.24& 9.12 & \multicolumn{1}{c}{9.93} & 12.73 & 9.76 \\ 

\multicolumn{1}{l|}{TENT} & \multicolumn{1}{c}{6.02} & \multicolumn{1}{c}{7.96} & 9.57 & 7.75 & \multicolumn{1}{c}{6.06} & 8.63  &6.71 & \multicolumn{1}{c}{9.91} &{10.26} & 10.55 & 12.33  &10.27 \\ 

\multicolumn{1}{l|}{EATA} & \multicolumn{1}{c}{6.05} & \multicolumn{1}{c}{7.96} & 9.74 & 7.93 & \multicolumn{1}{c}{6.06} &9.01 &6.24 &9.94 & 9.07 & \multicolumn{1}{c}{10.02} & 12.41 & 10.12 \\ 

\multicolumn{1}{l|}{MonoTTA} & \multicolumn{1}{c}{6.54} & \multicolumn{1}{c}{8.41} & 9.39 & 7.63 & \multicolumn{1}{c}{7.12} & 8.99 & 7.64 & \textbf{10.26} & 10.55 & \multicolumn{1}{c}{10.06} & 13.28 & 10.66 \\ 

\midrule
\multicolumn{1}{l|}{LearnableBN} & \multicolumn{1}{c}{\textbf{9.73}} & \multicolumn{1}{c}{\textbf{10.00}} & \textbf{9.65} & \textbf{10.16} & \multicolumn{1}{c}{\textbf{9.05}} &\textbf{10.85}& \textbf{8.09 }& 9.62 & \textbf{13.13}& \multicolumn{1}{c}{\textbf{14.74}} & \textbf{18.27} & \textbf{12.60} \\  

\bottomrule
\bottomrule

\end{tabular}
\caption{Comparison of different TTA methods on the \textbf{KITTI-C} validation set regarding Mean $AP_{3D|R_{40}}$ with IoU threshold set to 0.25 for the \textbf{Pedestrian} category. The baseline model is \textbf{Monoflex} . \textbf{Bold}: Best in the category.}

\label{Kitti-c}
\end{table*}

\subsection{Quantitative Results}

We compare our method to several test-time adaptation methods as shown in Table~\ref{baselinetable2}, 
These methods can be classified into two main categories, (1) adjusting model parameters based on unsupervised training (ie,TENT~\cite{wang2020tent}), (2) focuses on modifying the BN statistics (ie, ReviseBN~\cite{benz2021revisiting}, AdaBn~\cite{li2018adaptive}, ARM~\cite{zhang2021adaptive} )

The experimental results demonstrate that the BEV based model is highly sensitive to batch normalization (BN) statistics due to the number of parameters and model depth. ARM and AdaBn have caused the model to collapse. These methods have failed to predict the true distribution of the test domain, particularly when the batch size is minimal.
In response to this situation, the ReviseBN adjusts the mixture coefficient of the EMA method in accordance with the specific test domains.
ReviseBN showed significant improvements compared to the baseline in cases where the test domain greatly shifted from the training domain. For example, in the Motion blur corruption type, the average results improved from 0.3330 to 0.3555 compared to the baseline, However when the test domain was similar to the training domain, ReviseBN led to a degradation of the model's predictive ability. For example, the average results of the low light corruption type decreased from 0.3212 to 0.2631. 

TENT method fine-tunes the affine parameters of the BN layer using the EM loss. The TENT method has also caused a degradation in model performance, which is due to unsupervised training leading the model to fall into a local optimum. In severe domain shifts scenarios, TENT method is not as effective as the method of adjusting BN statistics.

Compared to these methods, our approach adaptively learns the mixture coefficients, by adaptively learning the mixture coefficients based on the different corruption scenarios and the varying depths of BN layers, which has overcome the instability issues commonly encountered in adjust BN statistics methods, and has effectively prevented model collapse.

The results of the experiments demonstrated that our method significantly enhanced the model's capacity for generalization in scenarios with severe domain shifts, including those involving fog, motion blur, and low light. Furthermore, as the degree of corruption increased, the efficacy of our method became increasingly evident. To illustrate, compared to the baseline in the low light corruption scenario, our method showed an improve the average performance from 0.3212 to 0.3469. Notably, in the hard severity, the performance improved significantly from 0.2274 to 0.2753.

At the same time, our method also demonstrated high stability in minimal domain shift scenarios. In the Fog corruption scenarios, our method avoids the performance degradation in model predictions that is commonly observed with common TTA methods. achieved the best average performance in the fog corruption scenarios.

To learn more about how LearnableBN helps models perform, we conducted experiments across ten different categories under snow corruption scenarios. The experimental results indicate that the introduction of the LearnableBN method did not result in a significant performance improvement when the baseline was already performing well. For instance, in the detection task for traffic cones, LearnableBN achieved only about a 14\% improvement compare to the baseline.
However, in the categories where the baseline model performs poorly, the LearnableBN method delivers significant improvements. Specifically, in the truck category, LearnableBN achieved a remarkable improvement of up to 342\% over the baseline, and in the pedestrian category, it enhanced performance by 283\%. It not only significantly enhances the model's robustness but also improves the generalization ability of autonomous driving models when faced with different categories of objects. These improvements are crucial for enhancing the reliability and safety of autonomous driving systems.
\begin{figure}
    \centering
    \includegraphics[width=1\linewidth]{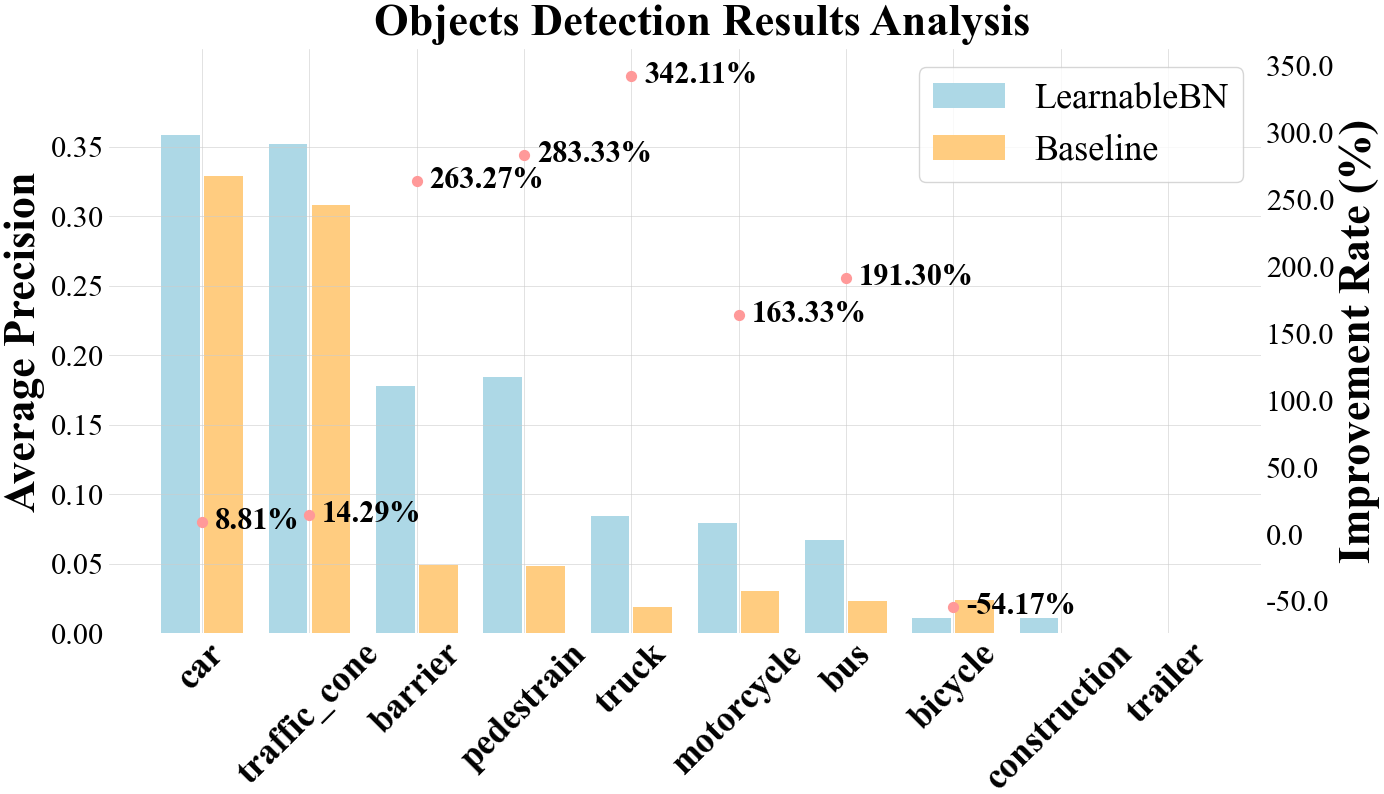}
    \caption{Comparison of Detection Results for Different Categories in Snow Scenarios and the severity is Hard. The baseline model is BEVFormer.}
    \label{different categories}
\end{figure}

\begin{table}[htbp] 
\centering
\small
\setlength{\tabcolsep}{9pt}
\begin{tabular}{l|ccc|c}
\toprule
 
\textbf{}&\multicolumn{4}{c}{\textbf{Motion Blur}}\\

\midrule
\multicolumn{1}{l|}{\textbf{Severity}} & \multicolumn{1}{c}{\textbf{Easy}} & \multicolumn{1}{c}{\textbf{Mid}} & \multicolumn{1}{c}{\textbf{Hard}} & \textbf{Avg}\\
\midrule
\multicolumn{1}{l|}{Sparse4D} & \multicolumn{1}{c}{0.4809} & \multicolumn{1}{c}{0.3189} & \multicolumn{1}{c}{0.269} & \multicolumn{1}{c}{0.3563}\\

\multicolumn{1}{l|}{TENT} & \multicolumn{1}{c}{\underline{0.4868}} & \multicolumn{1}{c}{0.368} & \multicolumn{1}{c}{0.3188} & \multicolumn{1}{c}{\underline{0.3912}}\\  

\multicolumn{1}{l|}{ReviseBN} & \multicolumn{1}{c}{0.4533} & \multicolumn{1}{c}{\underline{0.374}} & \multicolumn{1}{c}{\textbf{0.3359}} & \multicolumn{1}{c}{0.3877}\\

\multicolumn{1}{l|}{AdaBN} & \multicolumn{1}{c}{0.1713} & \multicolumn{1}{c}{0.1737} & \multicolumn{1}{c}{0.1442} & \multicolumn{1}{c}{0.1630}\\

\midrule
\multicolumn{1}{l|}{LearnableBN} & \multicolumn{1}{c}{\textbf{0.4962}} & \multicolumn{1}{c}{\textbf{0.3795}} & \multicolumn{1}{c}{\underline{0.335}} & \multicolumn{1}{c}{\textbf{0.4035}}\\  

\bottomrule
\end{tabular}
\caption{Comparison of different TTA methods across three levels of severity Motion Blur in \textbf{Sparse4D} with \textbf{ResNet-101} as the backbone. \textbf{Bold}: Best in the category. \underline{Underline}: Second best in the category.}
\label{sparse4D}
\end{table}

\subsection{Generalization Evaluation}
Additionally, we substituted the baseline model with the Sparse4D model and compared it with three representative TTA methods in the Motion Blur corruption scenario.
We selected Motion Blur for comparison because it presents significant domain shifts across the easy, mid and hard severity levels, which helps us assess our method's performance under both minimal and severe domain shifts.
As shown in Table~\ref{sparse4D}, The experimental results demonstrate that our method exhibits robust performance across all severity levels, with an average improvement in performance from 0.3563 to 0.4035 in comparison to the baseline. Consistent with the experiment results using BEVFormer as the baseline model, our method proves to be more stable than the methods that adjust BN statistics. 

On the other hand, to further validate the performance of the LearnableBN method in different real-world scenarios and tasks, we tested various TTA methods (BN adaptation~\cite{schneider2020improving}, TENT~\cite{wang2020tent}, EATA~\cite{niu2022efficient}, MonoTTA~\cite{lin2024fully}) on the KITTI-C dataset using the monocular 3D object detection task, we compared the experimental results presented in the MonoTTA paper~\cite{lin2024fully}. As shown in Table~\ref{Kitti-c}, the experimental results show that under real-world corruptions, the pre-trained model suffers from significant performance degradation due to data distribution shifts. The LearnableBN method brings a substantial average performance improvement on MonoFlex and maintains the best performance in detecting pedestrians in the KITTI-C dataset.

These experiments demonstrate that the LearnableBN method is adaptable to a wide range of base models, tasks, and real-world scenarios, highlighting its broad applicability and generalizability.

\begin{table*} [htbp] 
\centering
\small
\setlength{\tabcolsep}{8pt}
\begin{tabular}{ccc|cccccc}
\toprule
 
\textbf{LearnableBN}&\textbf{GSEM}&\textbf{Dual-stage}&\multicolumn{1}{c}{\textbf{Snow}} & \multicolumn{1}{c}{\textbf{Motion blur}} & \multicolumn{1}{c}{\textbf{Brightness}}& \multicolumn{1}{c}{\textbf{Low Light}} & \multicolumn{1}{c}{\textbf{Fog}} & \multicolumn{1}{c}{\textbf{Color Quant}}\\

\midrule
&  &  & \multicolumn{1}{c}{{0.2297}} & \multicolumn{1}{c}{{0.3330}} &\multicolumn{1}{c}{{0.4908}} &\multicolumn{1}{c}{{0.3212}} &\multicolumn{1}{c}{{0.4796}} &\multicolumn{1}{c}{{0.4184}}\\

\checkmark &  &  & \multicolumn{1}{c}{{0.2275}} & \multicolumn{1}{c}{{0.3289}} &\multicolumn{1}{c}{{0.4575}} &\multicolumn{1}{c}{{0.3195}} &\multicolumn{1}{c}{{0.4625}} &\multicolumn{1}{c}{{0.3991}}\\

\checkmark & \checkmark &  & \multicolumn{1}{c}{{0.2509}} & \multicolumn{1}{c}{{0.3712}} &\multicolumn{1}{c}{{0.4583}} &\multicolumn{1}{c}{{0.3473}} &\multicolumn{1}{c}{{0.4615}} &\multicolumn{1}{c}{{0.4004}}\\

\checkmark & \checkmark & \checkmark & \multicolumn{1}{c}{{0.2639}} & \multicolumn{1}{c}{{0.3787}} &\multicolumn{1}{c}{{0.4835}} &\multicolumn{1}{c}{{0.3469}} &\multicolumn{1}{c}{{0.4808}} &\multicolumn{1}{c}{{0.4192}}\\

\bottomrule
\end{tabular}

\caption{Ablation study of our method with LearnableBN, GSEM (Generalized-search Entropy Minimization) and Dual-stage (Semantic-Consistency based Dual-Stage-Adaptation). Comparison of the \textbf{average performance} across three levels of severity for six types of corruption. Since the secondary correction is not a independent component, it is included in the GSEM.}
\label{ablation}
\end{table*}

\subsection{Ablation Studies}
\noindent\textbf{LearnableBN.}~As shown in Table~\ref{ablation}, applying LearnableBN to the baseline results in degraded performance. 
This degradation is due to the secondary correction is not in this component resulting in the learned mixture coefficients not being able to adjust the BN statistic in time at each training step, which leads to a degradation of the model's performance.
It can be observed that after the LearnableBN method was applied to the baseline, the model's performance remained at a similar level to that of the baseline, avoiding the instability that is typically caused by adjusting BN statistics.

\noindent\textbf{Generalized-search Entropy Minimization.}~Compared to LearnableBN component, the GSEM component modifies the EM loss with GSEM loss and introduces a secondary correction step in the training process. As shown in Table~\ref{ablation}, applying both LearnableBN and GSEM to the baseline significantly improves performance in snow, motion blur and low light corruption scenarios, indicating that GSEM and LearnableBN component have both improved performance.
However, the experimental result also demonstrates that even with the introduction of the GSEM component, the performance degradation in the brightness corruption scenario remains unresolved (performance declined from 0.4908 to 0.4583). This is due to the challenges of learning from unstable samples.

\noindent\textbf{Semantic-Consistency based Dual-Stage-Adaptation.} After introducing the semantic-consistency based dual-stage-adaptation method, compared to the results of only applying LearnableBN and GSEM to the baseline, we resolved the degradation in the brightness and fog corruption scenarios (improving performance from 0.4583 to 0.4835). This improvement is attributed to the dual-stage training, which filtered out unstable samples and further enhanced the stability of the training process. Additionally, performance improvements were also observed in the snow, motion blur, and color quant corruption scenarios. This is attributed to the adjustment of the learning rate during the dual-stage-adaptation, which encourages the model to converge to a globally optimal solution.

\begin{figure}
    \centering
    \begin{subfigure}{0.23\textwidth}
        \centering
        \includegraphics[width=\textwidth]{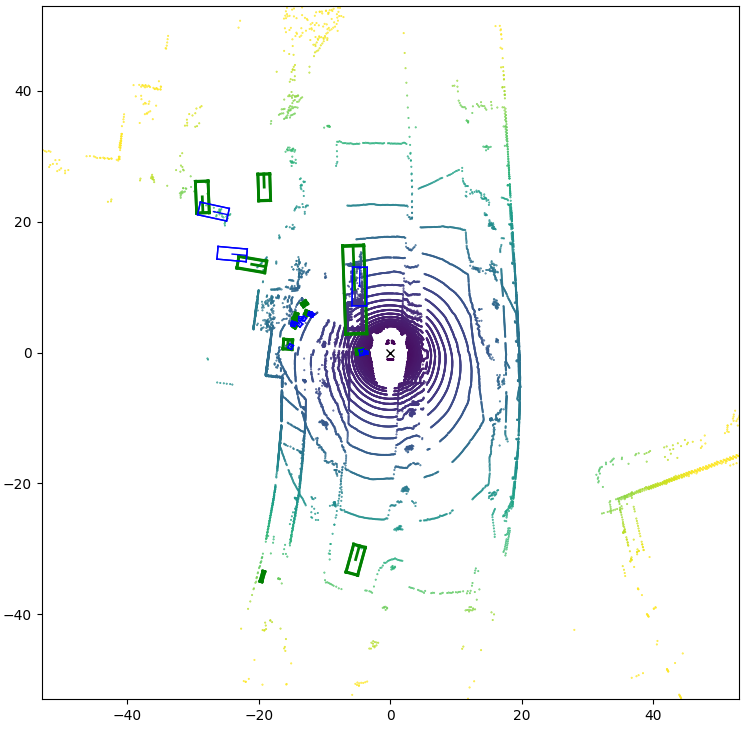} 
        \caption{BEVformer}
        \label{before_bev}
    \end{subfigure}
\hfill
    \begin{subfigure}{0.23\textwidth}
        \centering
        \includegraphics[width=\textwidth]{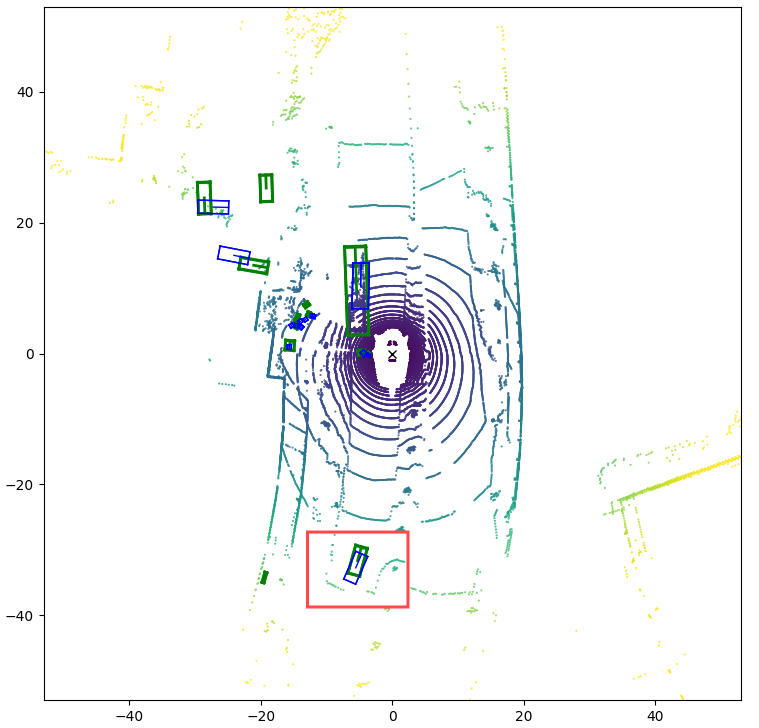}  
        \caption{LearnableBN}
        \label{after_bev}
    \end{subfigure}
    \caption{
    Example of BEV visualization results, where green bounding box is the ground truth, blue bounding box is the prediction results, and the red boxes highlight the difference before and after using our proposed LearnableBN.}
    \label{bev}
\end{figure}

\begin{figure}
    \centering
    \begin{subfigure}{0.45\textwidth}
        \centering
        \includegraphics[width=\textwidth]{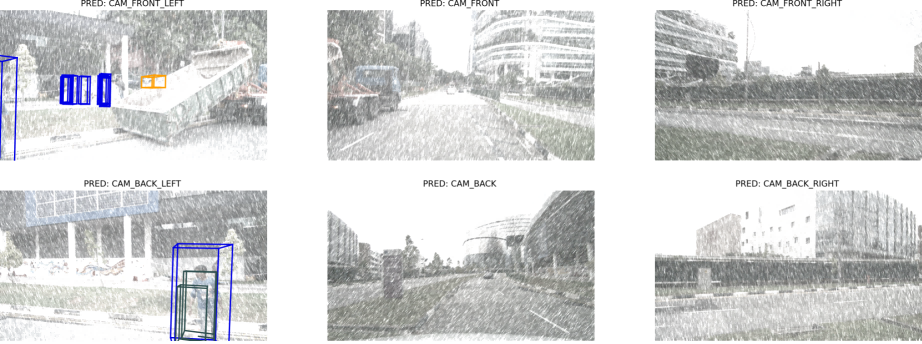} 
        \caption{Bevformer}
        \label{before_six}
    \end{subfigure}
\hfill
    \begin{subfigure}{0.45\textwidth}
        \centering
        \includegraphics[width=\textwidth]{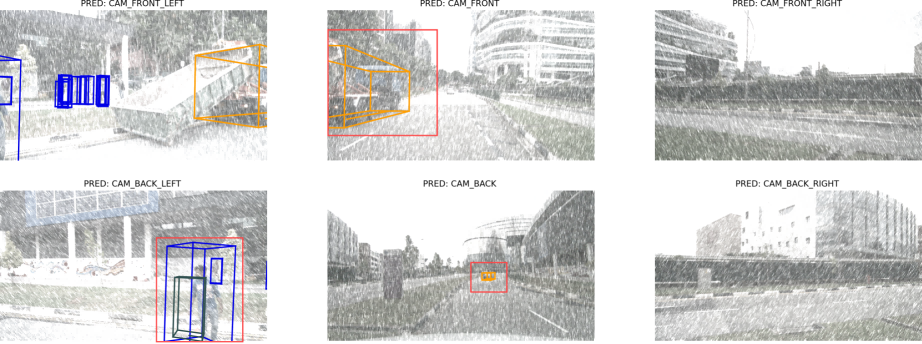}  
        \caption{LearnableBN}
        \label{after_six}
    \end{subfigure}
    \caption{Examples of visualization results w.r.t six perspectives, where the red boxes highlight the difference before and after using our proposed LearnableBN. Results on cars are colored in yellow, pedestrian in blue and cyclists in gleen.}
    \label{six}
\end{figure}

\subsection{Quantitative Results}

To verify the effectiveness of our LearnableBN method, we utilized BEVFormer as the baseline and focused on snow scenario for visualization analysis. Fig.~\ref{bev} presents the detection results in the BEV perspective, where Fig.~\ref{before_bev} shows results of BEVFormer, while Fig.~\ref{after_bev} shows results of BEVFormer after applying the LearnableBN method. It is clear from the figure that more objects can be detected after applying our proposed LearnableBN method. 

Furthermore, Fig.~\ref{six} provides visualization results from six different perspectives, where Fig.~\ref{before_six} represents the detection results of BEVFormer, and Fig.~\ref{after_six} illustrates the results of BEVFormer after applying LearnableBN. It can be observed that extreme weather conditions significantly degrade the detection ability of BEVFormer. By applying LearnableBN, not only were more objects detected, but also erroneous predictions were corrected compared to the baseline.

\section{Conclusion}
In this paper, we presented a LearnableBN to improve the robustness of perception models in real-world autonomous driving, which introduced auxiliary learnable parameters to the BN layer, and adopted the GSEM loss function. Additionally, we employed the semantic-consistency based dual-stage-adaptation to enhance generalization. Comprehensive experimental results demonstrated the effectiveness and superiority of our proposed methods.

{
    \small
    \bibliographystyle{ieeenat_fullname}
    \bibliography{main}
}

\input{sec/X_suppl}

\end{document}

%% file: preamble.tex
\usepackage[dvipsnames]{xcolor}

%% file: sec/X_suppl.tex
\clearpage
\setcounter{page}{1}
\maketitlesupplementary

\section{More method details}

The pseudo-code of Semantic-Consistency based Dual-Stage-Adaptation is summarized in Algorithm~\ref{algo}

\begin{algorithm}
\caption{\textbf{Semantic-Consistency based Dual-Stage-Adaptation}}
\label{algo}
\textbf{Input} $\{x_1,x_2,\dots,x_n\}$ in $D_c^s$, learning rate $\eta$,  auxiliary learnable parameter $\phi$ in BN layers, learning ratio $\alpha$, iteration $n_{max}$ and $m_{max}$\\
\textbf{Output} $f_{2}(x_n|\theta,\phi)$\\
\textbf{Stage 1:} Stable Adaptation
\begin{algorithmic}[1]
\STATE initialize $\eta=c$, $\phi=\epsilon$ {(a small positive constant)}
\FOR{ $n = 1, 2, \dots, n_{max}$} 
    \STATE $\mathcal{L}_{GSEM}=\mathcal{L}((f_{1}(x_n|\theta,\phi^{(n)}))$
    \STATE $\phi^{(n+1)}=\phi^{(n)}-\eta\cdot\nabla_{\phi}\mathcal{L}_{GSEM}$
    \STATE update BN statistics using $\phi^{(n+1)}$
\ENDFOR
\end{algorithmic}
\textbf{Stage 2:} Aggressive Adaptation
\begin{algorithmic}[1]
\addtocounter{ALC@line}{6}
\STATE initialize $\eta=C$ (C is larger than c);
\STATE $f_{2}({\cdot}|\theta,\phi)=(f_{1}({\cdot}|\theta,\phi)$ 
\FOR{ $m = 1, 2, \dots, m_{max}$} 
    \STATE $\mathcal{L}_{kl}=D_{kl}(f_{1}(x_m|\theta,\phi^{(n+m)})||f_{2}(x_m|\theta,\phi^{(n+m)}))$
    \STATE Add $\mathcal{L}_{kl}$ to list and find its position ratio $p$
    \IF{$p$ less than $\alpha$}
        \STATE $\mathcal{L}_{GSEM}=\mathcal{L}(f_{1}(x_m|\theta,\phi^{(n+m)}))$
        \STATE $\phi^{(n+m+1)}=\phi^{(n+m)}-\eta\cdot\nabla_{\phi}\mathcal{L}_{GSEM}$
        \STATE update BN statistics using $\phi^{(n+m+1)}$ 
    \ENDIF
\ENDFOR
\end{algorithmic}
\end{algorithm}

\begin{figure}
    \centering
    \includegraphics[width=1\linewidth]{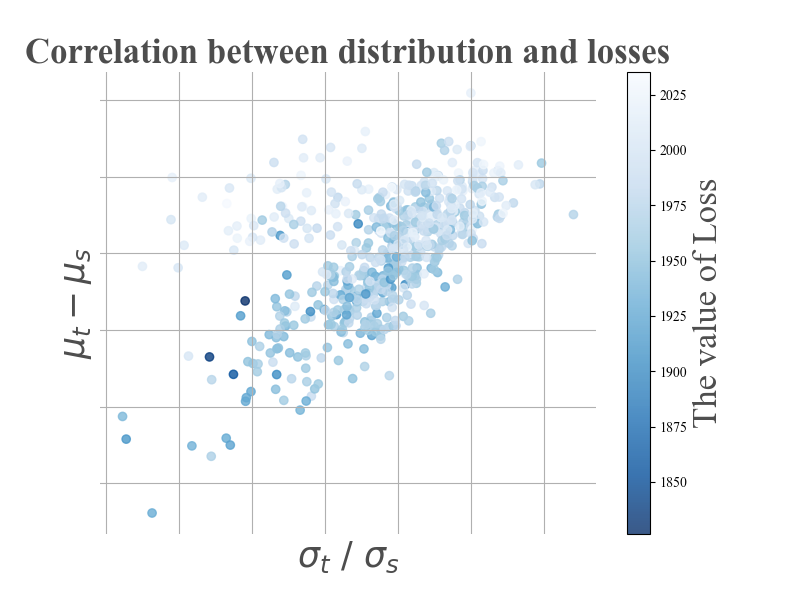}
    \caption{The distribution of samples, data points shift from deep blue to light blue as loss increases}
    \label{loss}
\end{figure}

\section{Data analysis}
\subsection {variation trend of $\phi$}
\begin{figure*}
    \centering
    \begin{subfigure}[b]{0.95\textwidth}
        \centering
        \caption{Snow}
        \includegraphics[width=\textwidth]{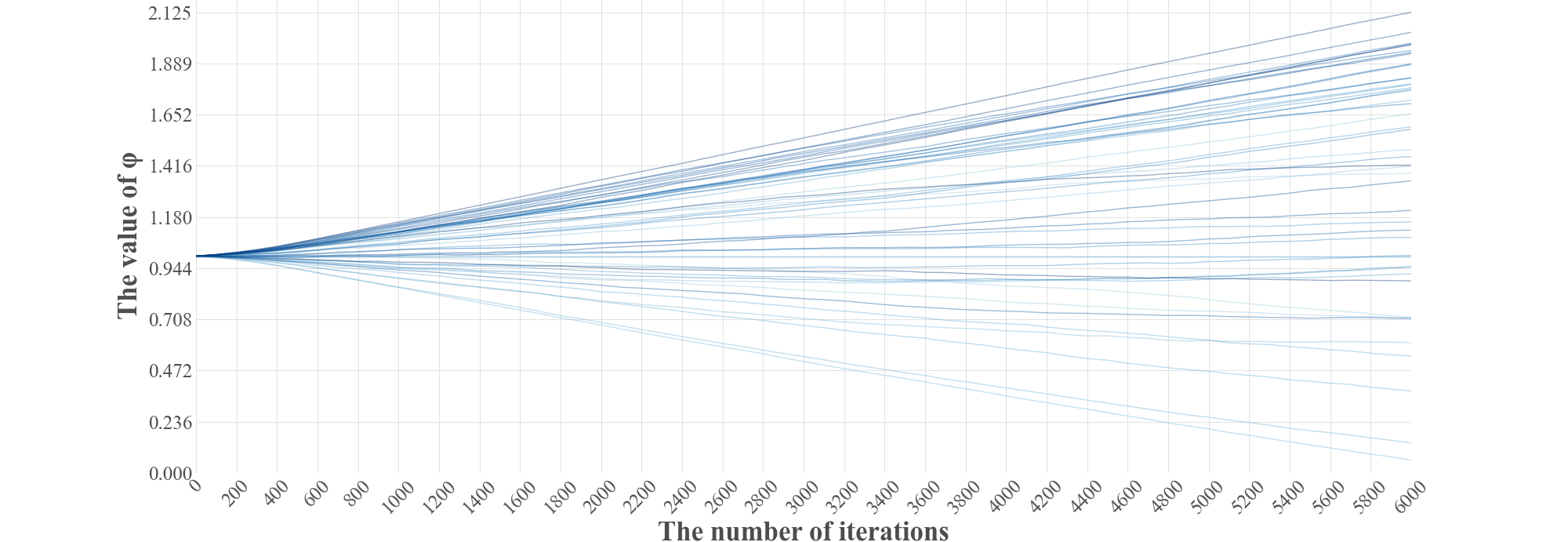} 
        \label{snow}
    \end{subfigure}
\vfill
    \begin{subfigure}[b]{0.95\textwidth}
        \centering
        \caption{Lowlight}
        \includegraphics[width=\textwidth]{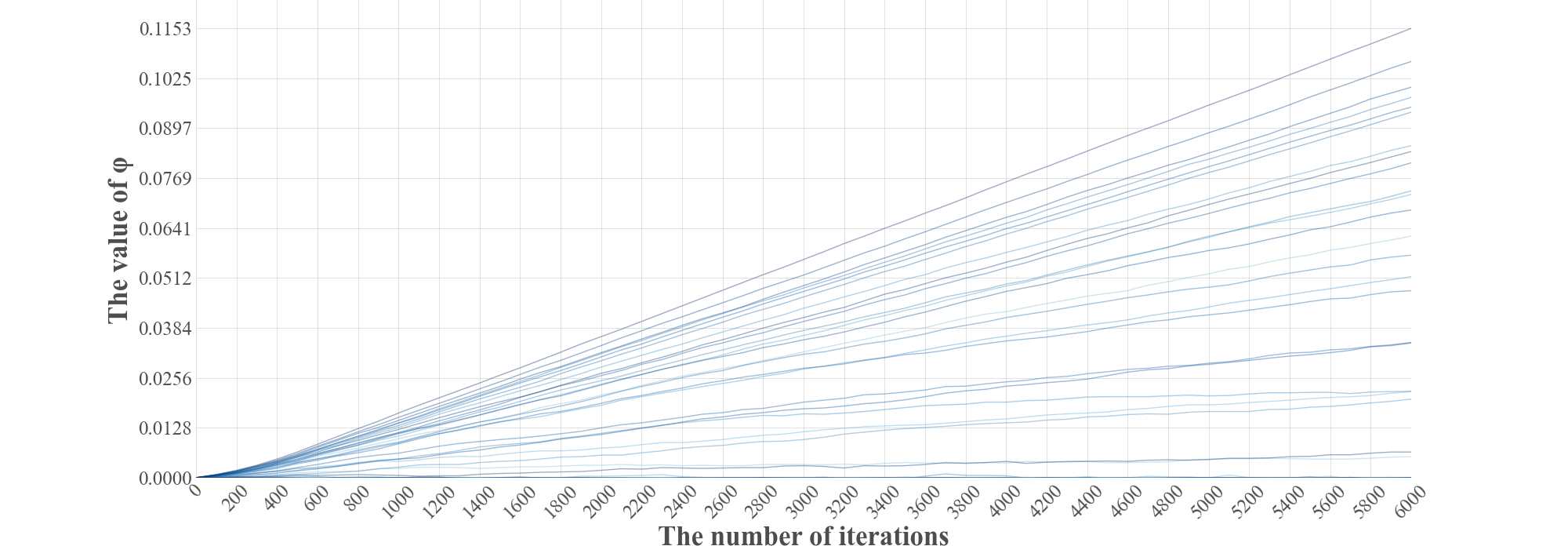}  
        \label{lowlight}
    \end{subfigure}
    \caption{Variation of $\phi$ values across different BN layers during model adapting in snow and lowlight scenarios. The deepening color of the lines corresponds to BN layers situated in the shallower regions of the model. In the snow scenario, the unit of the y-axis scale is $10^-5$.}
    \label{phi}
\end{figure*}

During training, we plotted the variations of the $\phi$ values in the BN layers, using different colors to distinguish the layers based on their depth within the model. As shown in the Fig~\ref{snow} and Fig~\ref{lowlight}. It can be observed that in the BN layers closer to the bottom of the model, the $\phi$ values continuously decrease, while in the BN layers closer to the output, the $\phi$ values continuously increase. This phenomenon indicates that the parameters in the bottom layers of the neural network have stronger transferability. Therefore, when fine-tuning BN layers, adopting a uniform transfer strategy across all layers may pose a risk of degrading model performance. It is advisable to minimize adjustments to the deeper layer parameters while increasing adjustments to the shallower layers.

we conducted a comparative analysis under snow and lowlight scenarios, we found that although the domain shift between the snow scenario and the training domain is greater than that in the lowlight scenario, the $\phi$ values obtained from training in the snow scenario are actually smaller, while the $\phi$ values obtained in the lowlight scenario are $10^5$ times greater than those in the snow scenario. This indicates that when rectify BN statistics, one should not rely on the data distribution at the pixel level but rather focus on the data distribution at the feature level. For scenarios with severe domain shifts, fine-tuning the BN statistics by an order of magnitude of $10^{-5}$ can significantly enhance model performance. In contrast, for scenarios with smaller domain shifts, even completely disregarding the training domain's statistics may not degrade model performance.

\subsection {The correlation between data distribution and loss}
Fig~\ref{loss} illustrates the distribution of the mean and variance among different samples. The horizontal axis represents the ratio of the test domain variance to the training domain variance, while the vertical axis represents the difference between the mean values of the test and training domains. The color of each point indicates the loss for each sample, with the color gradually lightening as the loss increases. From the Fig~\ref{loss}, it is evident that there is no significant relationship between the distribution of the samples and the magnitude of the loss. The shift in statistical metrics does not lead to an increase in the model's entropy minimization loss. Thus, filtering noisy samples based on entropy minimization loss is not an effective method for estimating the BN statistics in the test domain. Therefore, this paper chooses to use model consistency analysis to filter noisy samples.

\section{More Experiment details}

\subsection{Severity setting details}
The three Severity settings for each corruption are consistent with the severity settings used in Nuscenes-C, as shown in the Table \ref{severity}.

\begin{table*}[htbp] 
\centering
\small
\setlength{\tabcolsep}{4pt}
\begin{tabular}{c|c|c|c|c}
\toprule
 
\multicolumn{1}{l|}{} &\multicolumn{1}{c|}{\textbf{Parameter}} & \multicolumn{1}{c|}{\textbf{Easy}} & \multicolumn{1}{c|}{\textbf{Mid}} & \multicolumn{1}{c}{\textbf{Hard}}\\

\midrule
Bright & adjustment in HSV space & 0.2&0.4&0.5\\
\midrule
Dark & scale factor & 0.5&0.4&0.3\\
\midrule
Fog & (thickness,smoothness) & (2.0,2.0)&(2.5,1.5)&(3.0,1.4)\\
\midrule
Snow & \multicolumn{1}{p{4cm}|}{(mean,std, scale, threshold, blur radius, blur std, blending ratio)} & (0.1, 0,3, 3.0, 0.5, 10.0, 4.0, 0.8)&(0.2, 0.3, 2, 0.5, 12, 4, 0.7)&(0.55, 0.3, 4, 0.9, 12, 8, 0.7)\\
\midrule
Motion Blur & (radius,sigma) & (15,5)&(15,12)&(20,15)\\
\midrule
Color Quant & bit number & 5&4&3\\

\bottomrule
\end{tabular}
\caption{Detailed parameters uses for generating multi-level corruptions}
\label{severity}
\end{table*}

\subsection{More Metrics details}

We evaluate the performance of our method with the official nuScenes mertric: nuScenes Detection Score (NDS), which calculating a weighted sum of mAP, mATE, mASE, mAOE, mAVE, and mAAE. The first step is to convert TP errors into TP scores using the eq. \ref{mertric} :
 \begin{equation}
TP_{score}=max(1 - TP_{error}, 0.0)
\label{mertric}
\end{equation}
Then, a weight of 5 is assigned to mAP, and a weight of 1 is assigned to each of the TP scores, followed by calculating the normalized sum.

\begin{table*}[htbp] 
\centering
\small
\setlength{\tabcolsep}{6pt}
\begin{tabular}{l|c|c|cc|c|c|cc|c|c|cc}
\toprule
\toprule
 
\multicolumn{1}{l|}{\textbf{}}&\multicolumn{4}{c|}{\textbf{Snow}} & \multicolumn{4}{c|}{\textbf{Color Quant}} & \multicolumn{4}{c}{\textbf{Brightness}} \\

\midrule
\multicolumn{1}{l|}{\textbf{Severity}} & \multicolumn{1}{c}{\textbf{Easy}} & \multicolumn{1}{c}{\textbf{Mid}} & \multicolumn{1}{c}{\textbf{Hard}} & \textbf{Avg} & \multicolumn{1}{c}{\textbf{Easy}} & \multicolumn{1}{c}{\textbf{Mid}} & \multicolumn{1}{c}{\textbf{Hard}} & \textbf{Avg} & \multicolumn{1}{c}{\textbf{Easy}} & \multicolumn{1}{c}{\textbf{Mid}} & \multicolumn{1}{c}{\textbf{Hard}} & \textbf{Avg} \\
%\midrule
%\multicolumn{13}{|c|}{This is a merged row across all columns} \\ 
\midrule
 
\multicolumn{1}{l|}{\textit{Baseline}} & \multicolumn{1}{c}{0.2864} & \multicolumn{1}{c}{0.2124} & 0.1905 & 0.2297 & \multicolumn{1}{c}{\textbf{0.5019}} & \multicolumn{1}{c}{\textbf{0.4378}} & 0.3157 & \underline{0.4184} & \multicolumn{1}{c}{\textbf{0.5153}} & \multicolumn{1}{c}{\textbf{0.4922}} & \textbf{0.465} & \textbf{0.4908} \\ 
 
\multicolumn{1}{l|}{ReviseBN} & \multicolumn{1}{c}{\underline{0.3271}} & \multicolumn{1}{c}{\textbf{0.2694}} & \underline{0.2009} & \textbf{0.2658} & \multicolumn{1}{c}{0.4433} & \multicolumn{1}{c}{0.4132} & \underline{0.3223} & 0.3929 & \multicolumn{1}{c}{0.4515} & \multicolumn{1}{c}{0.4428} & 0.4295 & 0.4412 \\ 
 
\multicolumn{1}{l|}{TENT} & \multicolumn{1}{c}{0.1857} & \multicolumn{1}{c}{0.146} & 0.1348 & 0.1555& \multicolumn{1}{c}{0.3518} & \multicolumn{1}{c}{0.2956} & 0.1939 & 0.2804 & \multicolumn{1}{c}{0.3625} & \multicolumn{1}{c}{0.3437} & 0.3217 & 0.3426 \\

\multicolumn{1}{l|}{AdaBn} & \multicolumn{1}{c}{0.1045} & \multicolumn{1}{c}{0.0478} & 0.061 & 0.0711 & \multicolumn{1}{c}{0.1367} & \multicolumn{1}{c}{0.1774} & 0.0787 & 0.1309 & \multicolumn{1}{c}{0.0549} & \multicolumn{1}{c}{0.0506} & 0.0395 & 0.0483 \\

\multicolumn{1}{l|}{ARM(BN)} & \multicolumn{1}{c}{0.1201} & \multicolumn{1}{c}{0.0921} & 0.0946 & 0.1022 & \multicolumn{1}{c}{0.1541} & \multicolumn{1}{c}{0.1821} & 0.1182 & 0.1514  & \multicolumn{1}{c}{0.0943} & \multicolumn{1}{c}{0.0897} & 0.0825 & 0.0888 \\ 
 
\midrule
\multicolumn{1}{l|}{LearnableBN} & \multicolumn{1}{c}{\textbf{0.3325}} & \multicolumn{1}{c}{\underline{0.2579}} & \textbf{0.2015} & \underline{0.2639} & \multicolumn{1}{c}{\underline{0.4886}} & \multicolumn{1}{c}{\underline{0.4343}} & \textbf{0.3349} & \textbf{0.4192} & \multicolumn{1}{c}{\underline{0.5009}} & \multicolumn{1}{c}{\underline{0.4853}} &\underline{ 0.4643} & \underline{0.4835} \\ 
\bottomrule

\bottomrule

\end{tabular}
\caption{Comparison of different TTA methods across three levels of severity  in the first three types of corruption. The baseline model is \textbf{BEVFormer} with \textbf{ResNet-101} as the backbone.\textbf{Bold}: Best in the category. \underline{Underline}: Second best in the category.}
\label{baselinetable}
\end{table*}

\begin{table*}[htbp] 
\centering
\small
\setlength{\tabcolsep}{6pt}
\begin{tabular}{l|c|c|c|cc|c|ccc|ccc}
\toprule
\toprule
\multicolumn{1}{l|}{\textbf{}} & \multicolumn{3}{c|}{\textbf{Noise}} & \multicolumn{3}{c|}{\textbf{Blur}} & \multicolumn{3}{c|}{\textbf{Weather}}& \multicolumn{3}{c}{\textbf{Digital}} \\ 
\midrule
\multicolumn{1}{l|}{\textbf{Method}} & \multicolumn{1}{c}{\textbf{Gauss.}} & \multicolumn{1}{c}{\textbf{Shot}} &\multicolumn{1}{c|}{ \textbf{Impul.}} & \multicolumn{1}{c}{\textbf{Defoc.}} & \multicolumn{1}{c}{\textbf{Glass}} & \multicolumn{1}{c|}{\textbf{Motion}} & \multicolumn{1}{c}{\textbf{Frost}} & \multicolumn{1}{c}{\textbf{Fog}} & \multicolumn{1}{c|}{\textbf{Brit.}} & \multicolumn{1}{c}{\textbf{Contr.}} & \multicolumn{1}{c}{\textbf{Pixel.}} & \multicolumn{1}{c}{\textbf{Sat.}} \\ 
%\midrule
%\multicolumn{13}{|c|}{This is a merged row across all columns} \\ 
\midrule
 
\multicolumn{1}{l|}{\textit{Baseline}} & \multicolumn{1}{c}{3.84} & \multicolumn{1}{c}{7.48} & 5.31& 2.59& \multicolumn{1}{c}{3.73} & 11.05& 7.77& \multicolumn{1}{c}{7.57} & 24.87& \multicolumn{1}{c}{6.92} & 28.16&31.46\\ 
 
\multicolumn{1}{l|}{BN adaptation} & \multicolumn{1}{c}{13.58} & \multicolumn{1}{c}{21.93} & 18.78& 15.87& \multicolumn{1}{c}{8.59} & 24.32& 21.45& 24.63& 31.80& \multicolumn{1}{c}{30.58} & 41.04& 30.71\\ 

\multicolumn{1}{l|}{TENT} & \multicolumn{1}{c}{17.80} & \multicolumn{1}{c}{27.09} & 23.18& 21.66& \multicolumn{1}{c}{11.90} & 28.75&26.58& \multicolumn{1}{c}{30.78} &35.65& 34.72& 41.71&35.91\\ 

\multicolumn{1}{l|}{EATA} & \multicolumn{1}{c}{16.67} & \multicolumn{1}{c}{26.42} & 25.07& 22.54& \multicolumn{1}{c}{13.23} &27.73&26.58&31.10& 35.39& \multicolumn{1}{c}{35.28} & 41.40& 36.72\\ 

\multicolumn{1}{l|}{MonoTTA} & \multicolumn{1}{c}{\textbf{21.15}} & \multicolumn{1}{c}{\textbf{28.65}} & \textbf{26.64}&\textbf{ 25.91}& \multicolumn{1}{c}{\textbf{19.26}} & \textbf{31.48}& \textbf{30.24}& \textbf{33.75}& \textbf{36.84}& \multicolumn{1}{c}{\textbf{36.83}} & \textbf{41.97}& \textbf{38.13}\\ 

\midrule
\multicolumn{1}{l|}{LearnableBN} & \multicolumn{1}{c}{15.70} & \multicolumn{1}{c}{22.53} & 20.12& 17.18& \multicolumn{1}{c}{7.55} &18.04& 20.14& 23.79& 31.96& \multicolumn{1}{c}{30.06} & 40.90& 32.65\\  

\bottomrule
\bottomrule

\end{tabular}
\caption{Comparison of different TTA methods on the \textbf{KITTI-C} validation set regarding Mean $AP_{3D|R_{40}}$ with IoU threshold set to 0.5 for the \textbf{Car} category. The baseline model is \textbf{Monoflex} . \textbf{Bold}: Best in the category.}

\label{More Kitti-c}
\end{table*}

\begin{table*}[htbp] 
\centering
\small
\setlength{\tabcolsep}{6pt}
\begin{tabular}{l|c|c|c|cc|c|ccc|ccc}
\toprule
\toprule
\multicolumn{1}{l|}{\textbf{}} & \multicolumn{3}{c|}{\textbf{Noise}} & \multicolumn{3}{c|}{\textbf{Blur}} & \multicolumn{3}{c|}{\textbf{Weather}}& \multicolumn{3}{c}{\textbf{Digital}} \\ 
\midrule
\multicolumn{1}{l|}{\textbf{Method}} & \multicolumn{1}{c}{\textbf{Gauss.}} & \multicolumn{1}{c}{\textbf{Shot}} &\multicolumn{1}{c|}{ \textbf{Impul.}} & \multicolumn{1}{c}{\textbf{Defoc.}} & \multicolumn{1}{c}{\textbf{Glass}} & \multicolumn{1}{c|}{\textbf{Motion}} & \multicolumn{1}{c}{\textbf{Frost}} & \multicolumn{1}{c}{\textbf{Fog}} & \multicolumn{1}{c|}{\textbf{Brit.}} & \multicolumn{1}{c}{\textbf{Contr.}} & \multicolumn{1}{c}{\textbf{Pixel.}} & \multicolumn{1}{c}{\textbf{Sat.}} \\ 
%\midrule
%\multicolumn{13}{|c|}{This is a merged row across all columns} \\ 
\midrule
 
\multicolumn{1}{l|}{\textit{Baseline}} & \multicolumn{1}{c}{0.28} & \multicolumn{1}{c}{1.64} & 0.47& 0.59& \multicolumn{1}{c}{2.97} & 3.60& 7.42& \multicolumn{1}{c}{3.81} & 10.07& \multicolumn{1}{c}{3.79} & 3.80&8.39\\

\multicolumn{1}{l|}{MonoTTA} & \multicolumn{1}{c}{3.21} & \multicolumn{1}{c}{3.97} & 3.70& \textbf{5.26}& \multicolumn{1}{c}{3.96} & 6.83&  \textbf{7.60}& 7.48& 9.32& \multicolumn{1}{c}{8.76} & 11.16& 7.76\\ 

\midrule
\multicolumn{1}{l|}{LearnableBN} & \multicolumn{1}{c}{\textbf{3.60}} & \multicolumn{1}{c}{\textbf{5.36}} & \textbf{4.04}& 4.19& \multicolumn{1}{c}{\textbf{4.54}} &\textbf{7.05}& 5.84& \textbf{7.56}& \textbf{11.39}& \multicolumn{1}{c}{\textbf{9.51}} &\textbf{ 11.95}& \textbf{10.90}\\  

\bottomrule
\bottomrule

\end{tabular}
\caption{Comparison of different TTA methods on the \textbf{KITTI-C} validation set regarding Mean $AP_{3D|R_{40}}$ with IoU threshold set to 0.25 for the \textbf{Cyclist} category. The baseline model is \textbf{Monoflex} . \textbf{Bold}: Best in the category.}

\label{More Kitti}
\end{table*}

\subsection{More implementation details}
\noindent\textbf{Implementation Details.} In Kitti-C, to evaluate the stability of TTA methods, the batch size was set to 8. In the semantic-consistency-based dual-stage-adaptation, we set the learning rates $\eta$ to 1000 and 100 and learning ratio $\alpha$ set at 0.1. The initial value of auxiliary learnable parameters $\phi$ in BN layers is set to 0.1.
In contrast to BEV-based 3D object detection, monocular object detection task requires setting the initial $\phi$ to a relatively large value. This adjustment is necessary because the MonoFlex model demonstrates a significant loss of predictive capability when processing corrupted data, severely compromising entropy minimization and leading to erroneous optimization of model parameters during training. By appropriately configuring the initial $\phi$, we aim to restore the model's predictive capacity in its initial state, thereby ensuring the stability and reliability of the adapting process.

\subsection{More Nuscenes-C Results}

As shown in the table~\ref{baselinetable}, we present the results of LearnableBN compared with other TTA methods on three additional types of corruption in the Nuscenes-C dataset.

The experimental results are consistent across the other three types of corruption. It can be observed that previous TTA methods exhibit instability when applied to models with a large number of parameters. For instance, while the ReviseBN method achieves the best average performance across three levels of corruption severity in the snow scenario, it leads to significant performance degradation in the color quant and brightness scenarios. This is because ReviseBN is not well-suited for models with a large number of parameters and is primarily effective in scenarios where corruption causes severe degradation of model performance.

In contrast, our proposed method, LearnableBN, demonstrates the highest robustness across various types of corruption. LearnableBN maintains the model's performance when detection accuracy is only slightly affected by corruption. For example, in the Color Quant scenario, LearnableBN improves the baseline's average performance across three severity levels from 41.84\% to 41.92\%. Additionally, when corruption severely impacts detection performance, LearnableBN significantly restores the model's predictive ability. For instance, in the snow scenario, it improves the baseline's average performance across three severity levels from 22.97\% to 26.39\%.

\subsection{More Kitti-C}

As shown in the table~\ref{More Kitti-c} and table~\ref{More Kitti}, We present a comparison of detection results for the Car and Cyclist categories on the Kitti-C dataset between LearnableBN and previous TTA methods. It is worth noting that for the detection results of the Cyclist category, MonoTTA's results are inconsistent with those shown in the paper. We believe this discrepancy stems from issues with the Cyclist category in the Kitti-C dataset provided by MonoTTA. As a result, for the Cyclist category, we only compare MonoTTA with the baseline and our LearnableBN method.

In the Car category, LearnableBN does not achieve optimal performance, performing worse than TENT, EATA, and MonoTTA, but outperforming BN Adaptation. This discrepancy can be attributed to the fact that TENT, EATA, and MonoTTA are unsupervised training methods that optimize the affine parameters of the BN layer. In contrast, both LearnableBN and BN Adaptation focus on adjusting the BN layer statistics. The advantage of unsupervised methods lies in their ability to improve detection results for categories with high prediction confidence. However, for multi-class scenarios, methods that adjust BN layer statistics demonstrate greater efficacy.
As shown in Table~\ and Table~, LearnableBN achieves near-optimal results in the Cyclist and Pedestrian categories, indicating its effectiveness in mitigating the long-tail effect and exhibits greater robustness compared to unsupervised test-time adaptation methods, making it particularly advantageous in handling diverse category distributions.

In small-parameter models with relatively shallow architectures, the issue of internal covariate shift is less pronounced. Additionally, in the Kitti-C dataset, model performance degrades significantly under corruption. LearnableBN is designed to address the instability of large-parameter models during test-time adaptation, which explains why it does not exhibit a clear advantage in this specific task.

In models with a small number of parameters, the architecture is relatively shallow, and the issue of internal covariate shift is less severe. Additionally, in the Kitti-C scenarios, the detection performance of models experiences significant degradation. Since LearnableBN is specifically designed to address the instability of large-parameter models during test-time adaptation, it does not demonstrate a notable advantage in this particular task.

\subsection{More Quantitative Results}
\begin{figure*}
    \centering
    \subcaptionbox*{Brightness}{%  
        \begin{subfigure}[b]{0.25\textwidth}
            \centering
            \caption*{easy}
            \includegraphics[width=\textwidth]{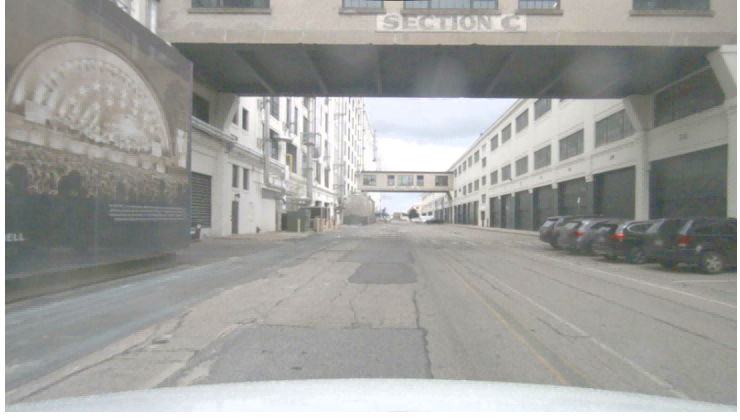}

        \end{subfigure}
        \hfill
        \begin{subfigure}[b]{0.25\textwidth}
            \centering
            \caption*{mid}
            \includegraphics[width=\textwidth]{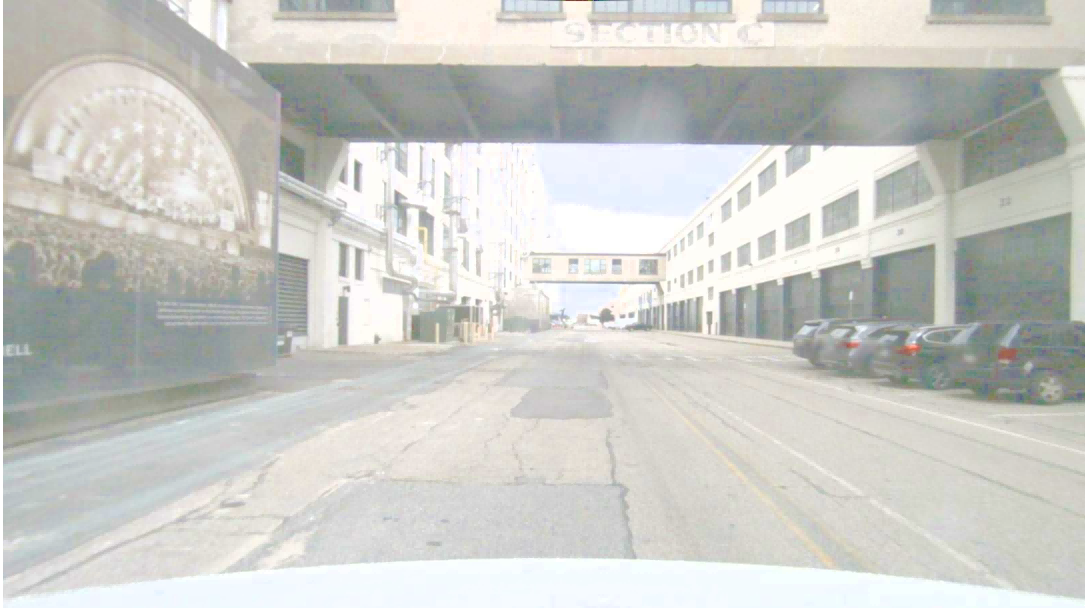}

        \end{subfigure}
        \hfill
        \begin{subfigure}[b]{0.25\textwidth}
            \centering
            \caption*{hard}
            \includegraphics[width=\textwidth]{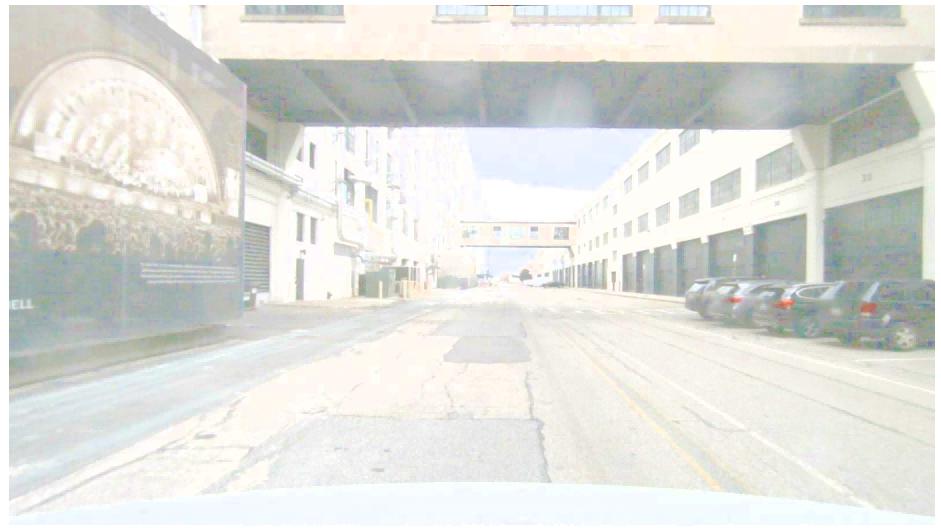}

        \end{subfigure}
    }
    \hfill
    \subcaptionbox*{Color Quant}{%  
        \begin{subfigure}[b]{0.25\textwidth}
            \centering
            \caption*{easy}
            \includegraphics[width=\textwidth]{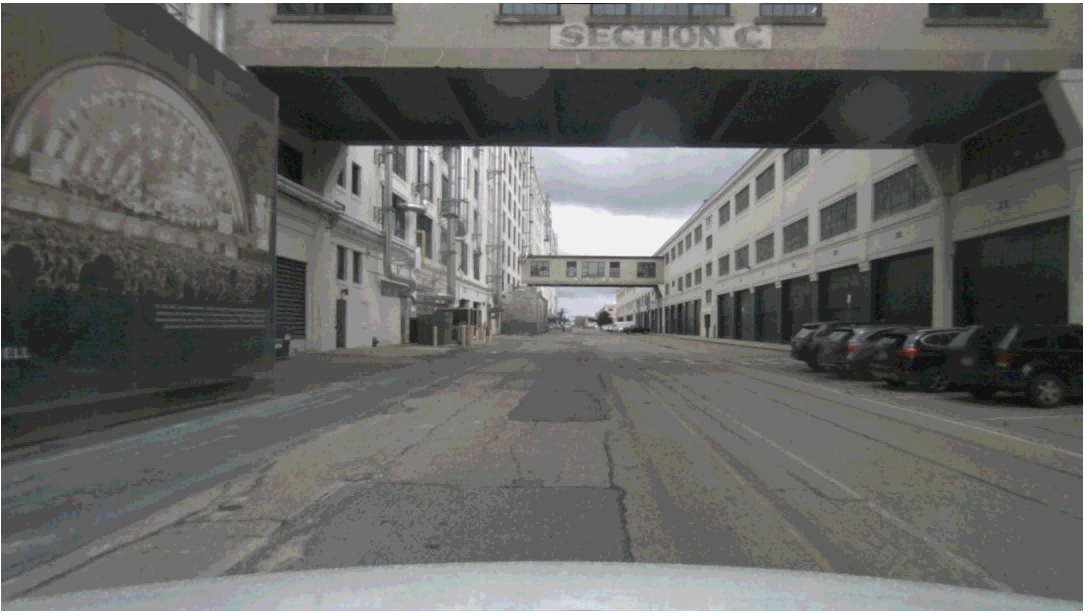}

        \end{subfigure}
        \hfill
        \begin{subfigure}[b]{0.25\textwidth}
            \centering
            \caption*{mid}
            \includegraphics[width=\textwidth]{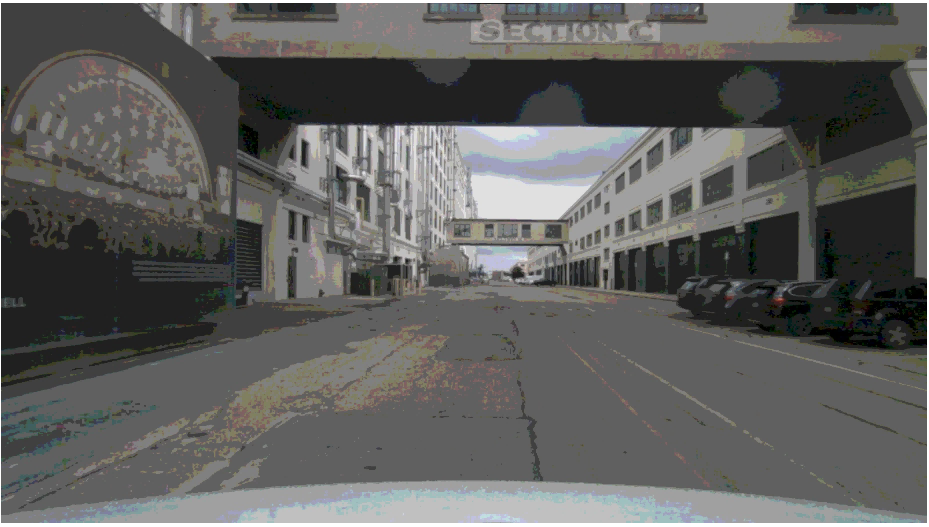}

        \end{subfigure}
        \hfill
        \begin{subfigure}[b]{0.25\textwidth}
            \centering
            \caption*{hard}
            \includegraphics[width=\textwidth]{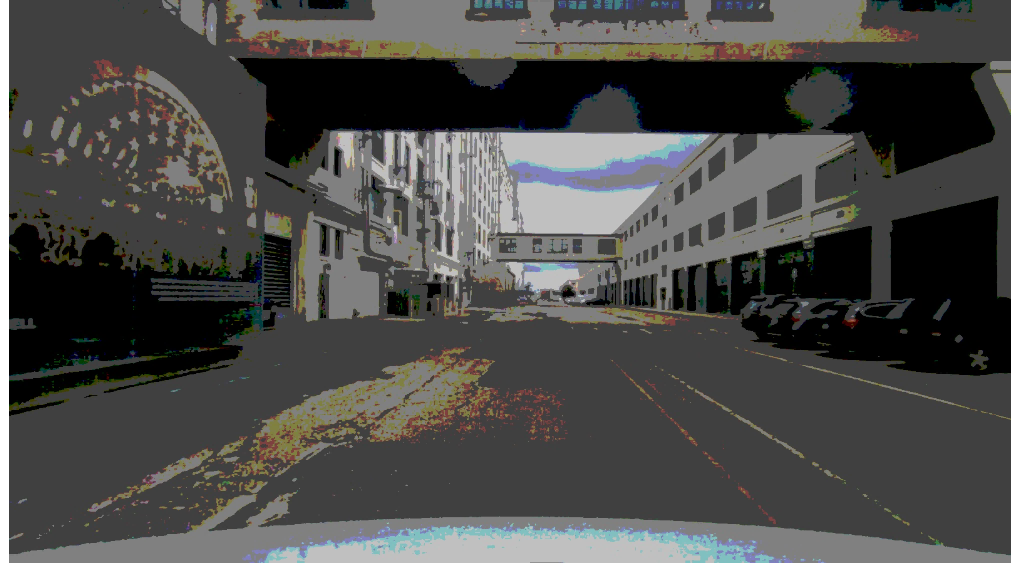}

        \end{subfigure}
    }
    \subcaptionbox*{Fog}{%  
        \begin{subfigure}[b]{0.25\textwidth}
            \centering
            \caption*{easy}
            \includegraphics[width=\textwidth]{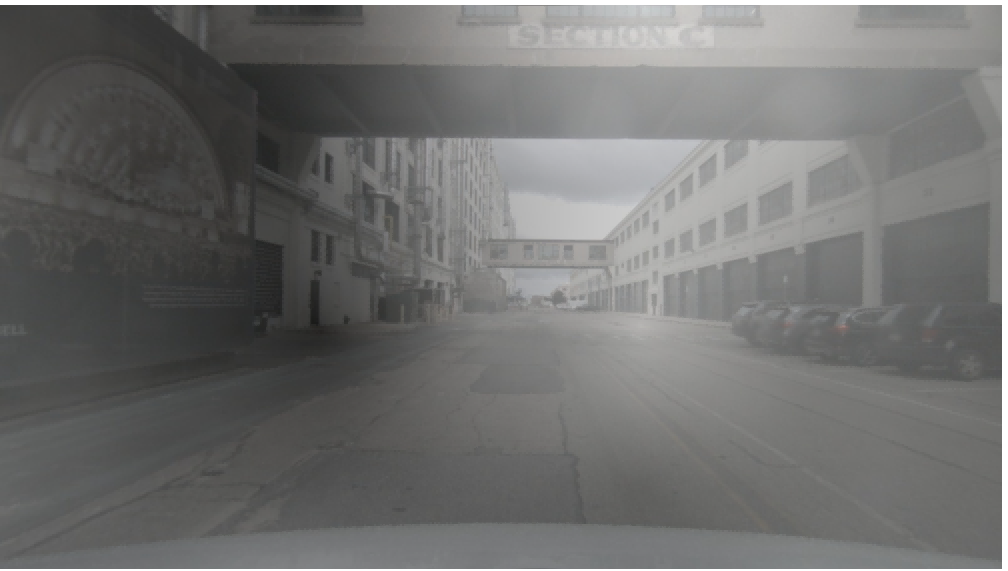}

        \end{subfigure}
        \hfill
        \begin{subfigure}[b]{0.25\textwidth}
            \centering
            \caption*{mid}
            \includegraphics[width=\textwidth]{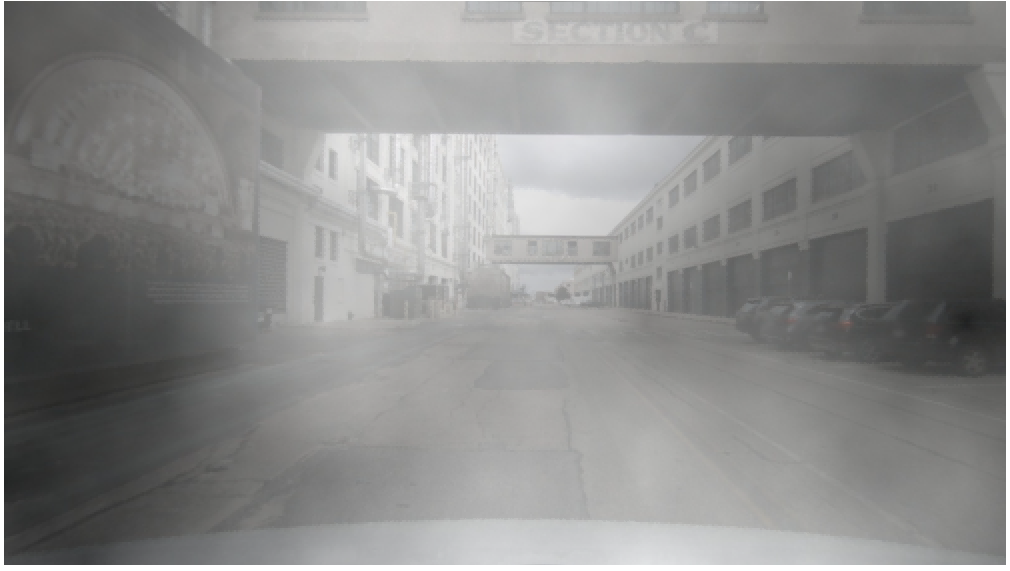}

        \end{subfigure}
        \hfill
        \begin{subfigure}[b]{0.25\textwidth}
            \centering
            \caption*{hard}
            \includegraphics[width=\textwidth]{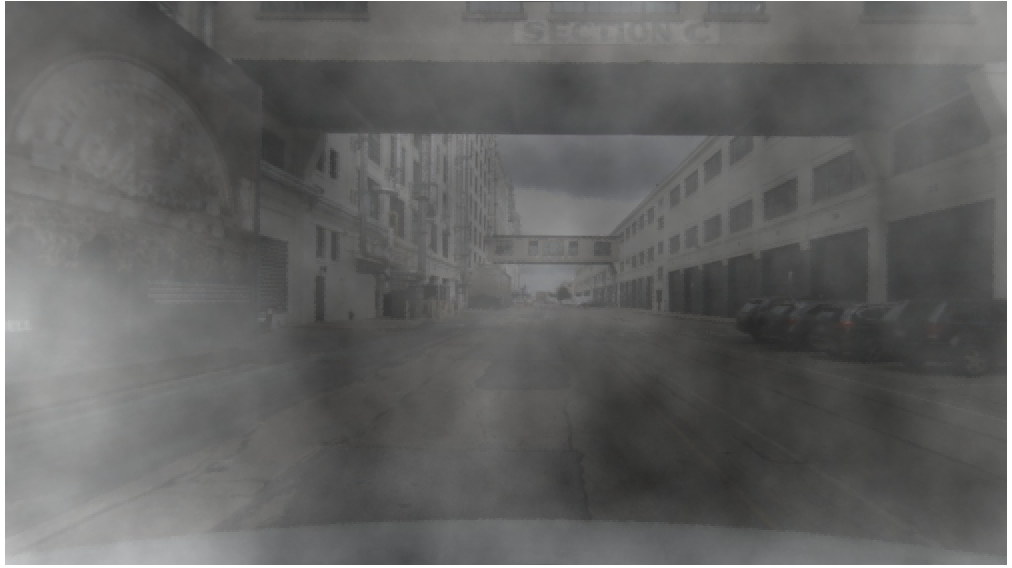}

        \end{subfigure}
    }
    \subcaptionbox*{Low Light}{%  
        \begin{subfigure}[b]{0.25\textwidth}
            \centering
            \caption*{easy}
            \includegraphics[width=\textwidth]{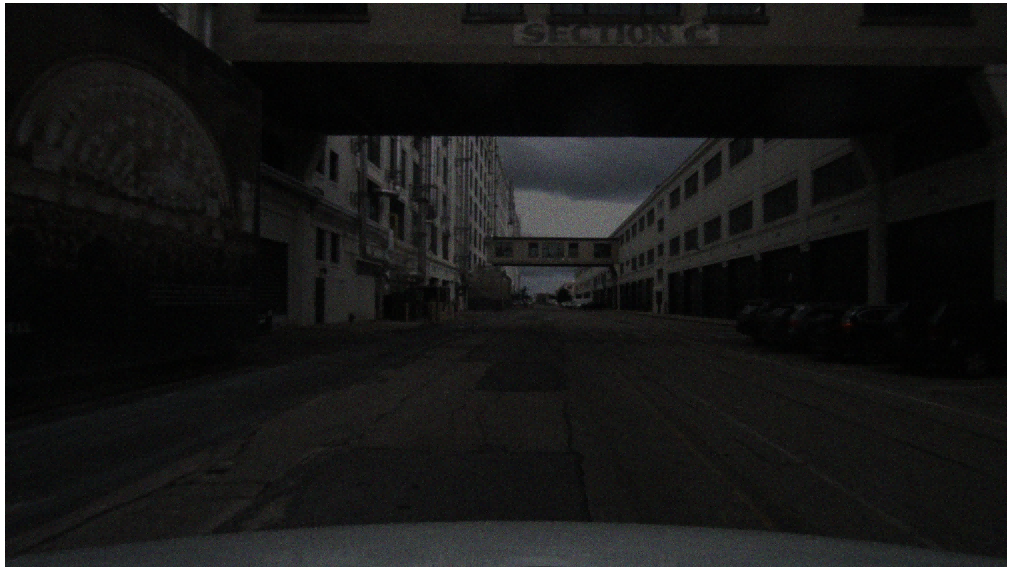}

        \end{subfigure}
        \hfill
        \begin{subfigure}[b]{0.25\textwidth}
            \centering
            \caption*{mid}
            \includegraphics[width=\textwidth]{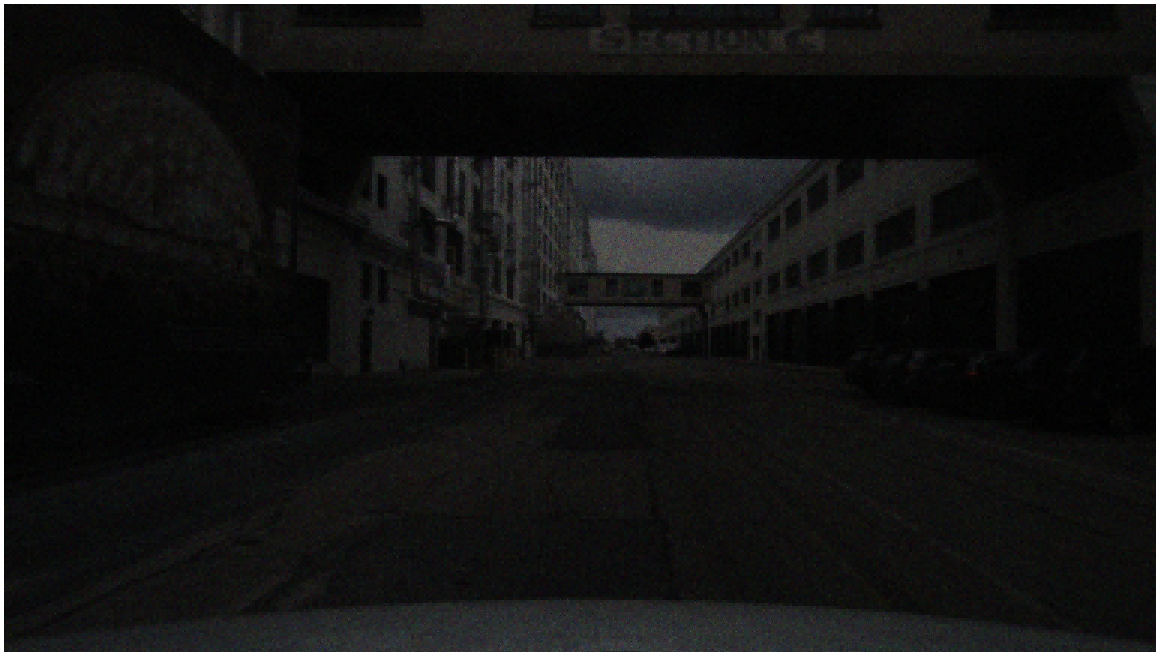}

        \end{subfigure}
        \hfill
        \begin{subfigure}[b]{0.25\textwidth}
            \centering
            \caption*{hard}
            \includegraphics[width=\textwidth]{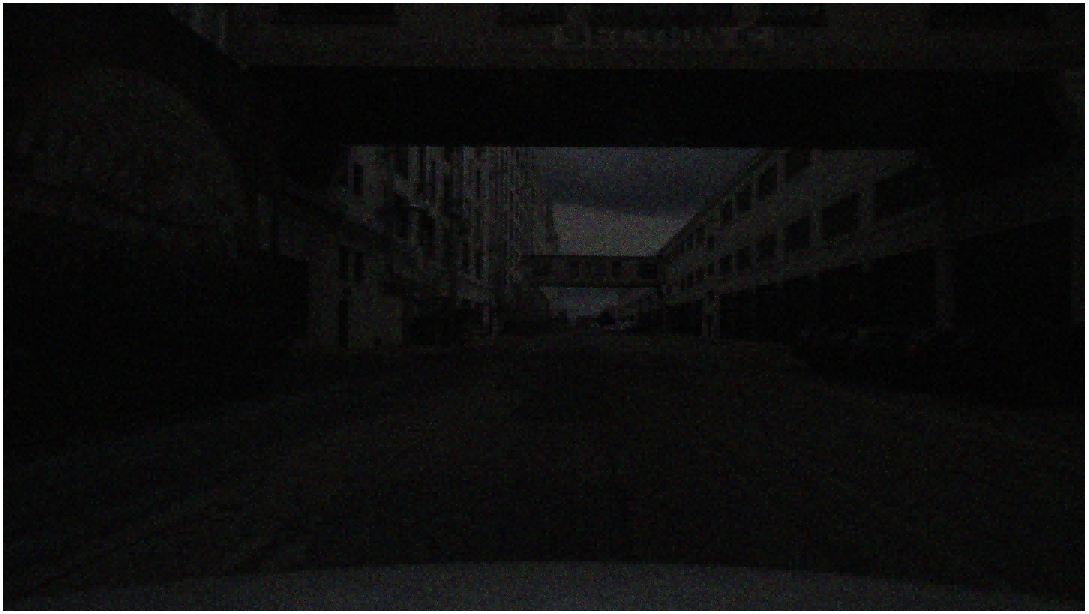}

        \end{subfigure}
    }
    \subcaptionbox*{Snow}{%  
        \begin{subfigure}[b]{0.25\textwidth}
            \centering
            \caption*{easy}
            \includegraphics[width=\textwidth]{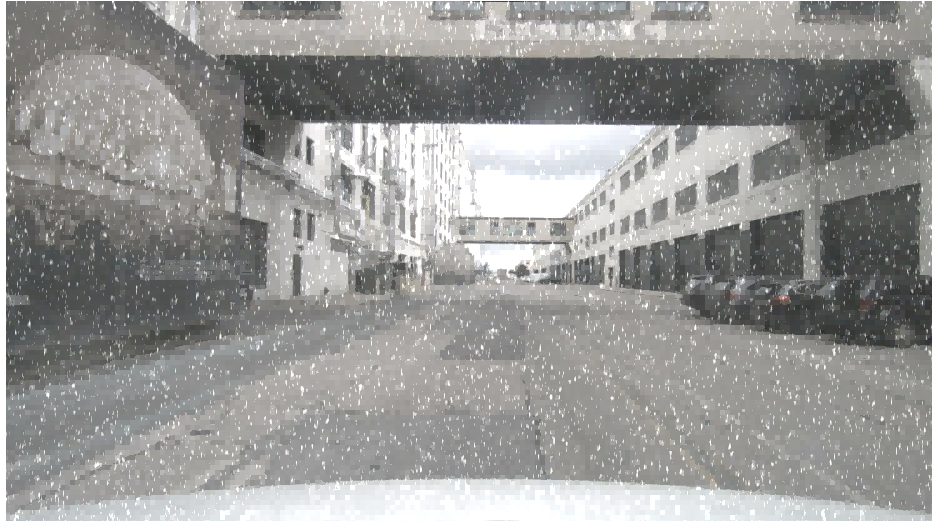}

        \end{subfigure}
        \hfill
        \begin{subfigure}[b]{0.25\textwidth}
            \centering
            \caption*{mid}
            \includegraphics[width=\textwidth]{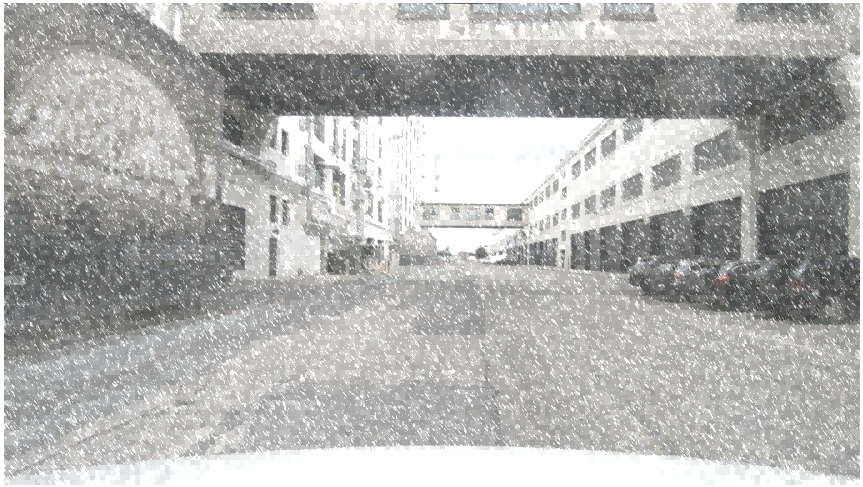}

        \end{subfigure}
        \hfill
        \begin{subfigure}[b]{0.25\textwidth}
            \centering
            \caption*{hard}
            \includegraphics[width=\textwidth]{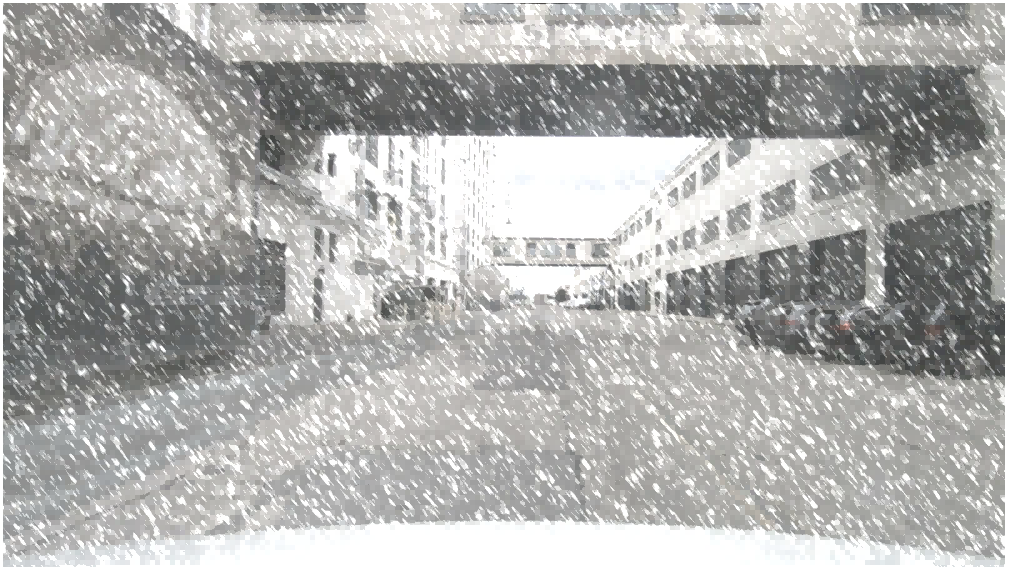}

        \end{subfigure}
    }
    \subcaptionbox*{Motion Blur}{%  
        \begin{subfigure}[b]{0.25\textwidth}
            \centering
            \caption*{easy}
            \includegraphics[width=\textwidth]{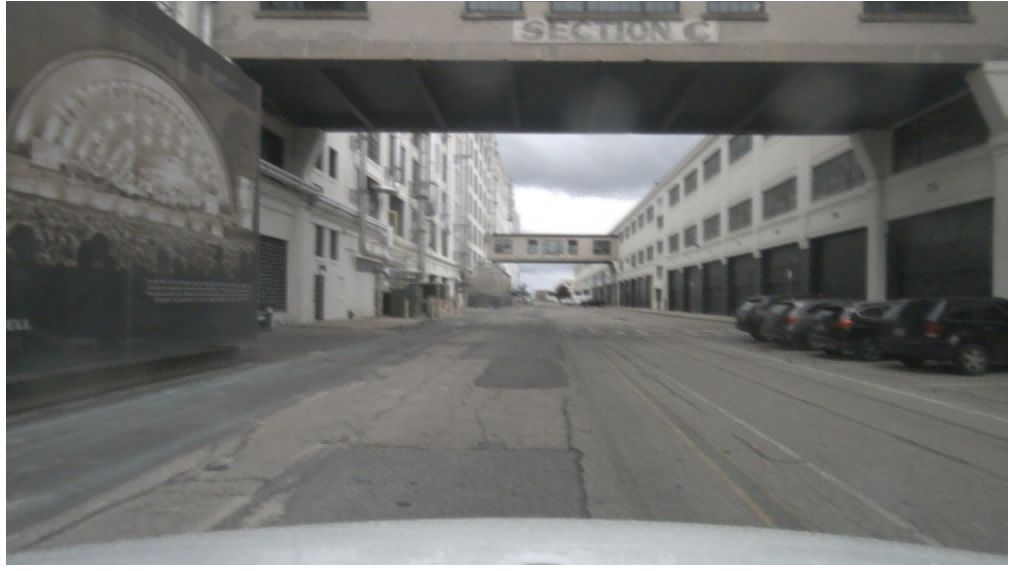}

        \end{subfigure}
        \hfill
        \begin{subfigure}[b]{0.25\textwidth}
            \centering
            \caption*{mid}
            \includegraphics[width=\textwidth]{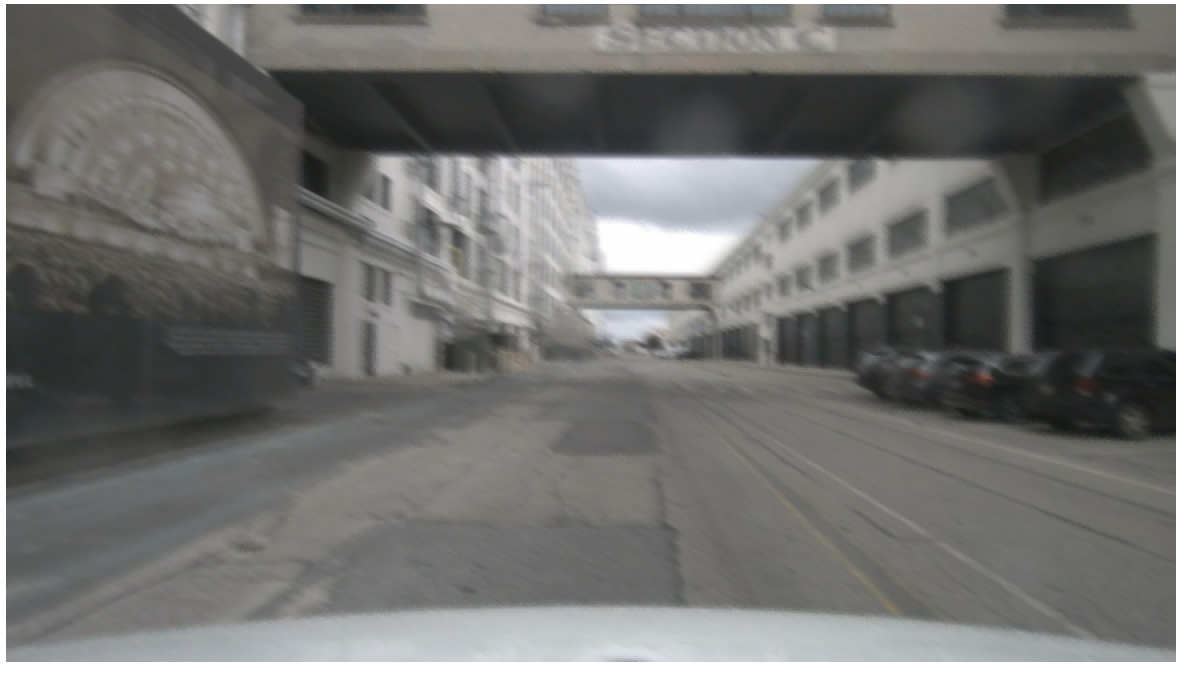}

        \end{subfigure}
        \hfill
        \begin{subfigure}[b]{0.25\textwidth}
            \centering
            \caption*{hard}
            \includegraphics[width=\textwidth]{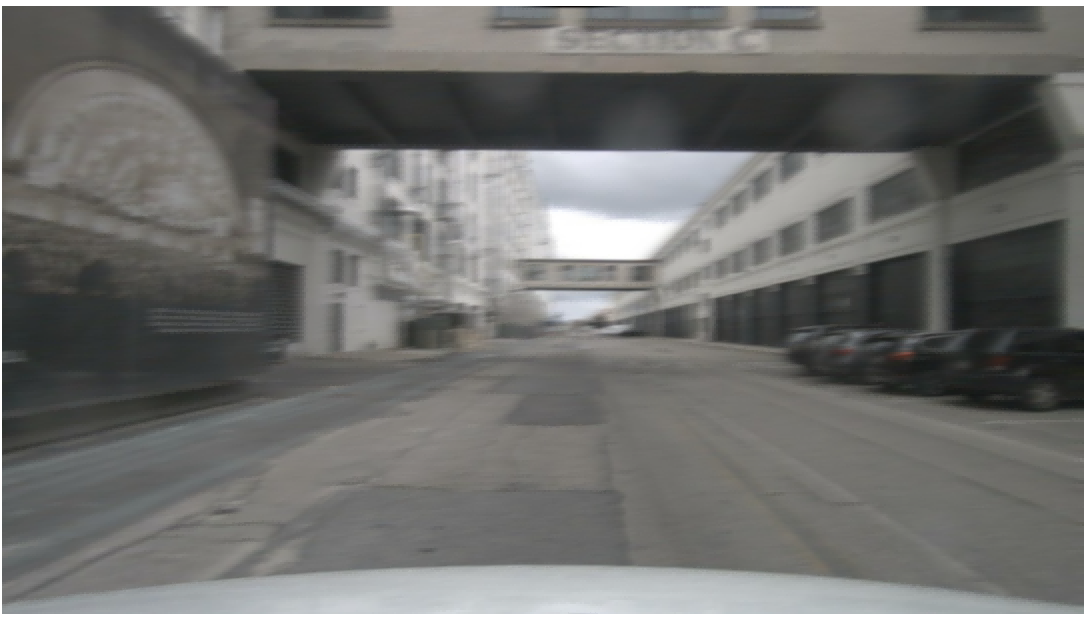}

        \end{subfigure}
    }
    \caption{the six distinct types of corruptions across three severity in Nuscenes-C dataset}
    \label{Nuscenes-c vis}
\end{figure*}

\begin{figure*}
    \centering
    \begin{subfigure}[b]{0.3\textwidth}
        \centering
        \includegraphics[width=\linewidth]{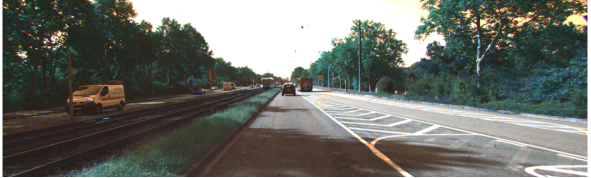}
        \caption*{Brightness}
        
    \end{subfigure}
    \hfill
    \begin{subfigure}[b]{0.3\textwidth}
        \centering
        \includegraphics[width=\linewidth]{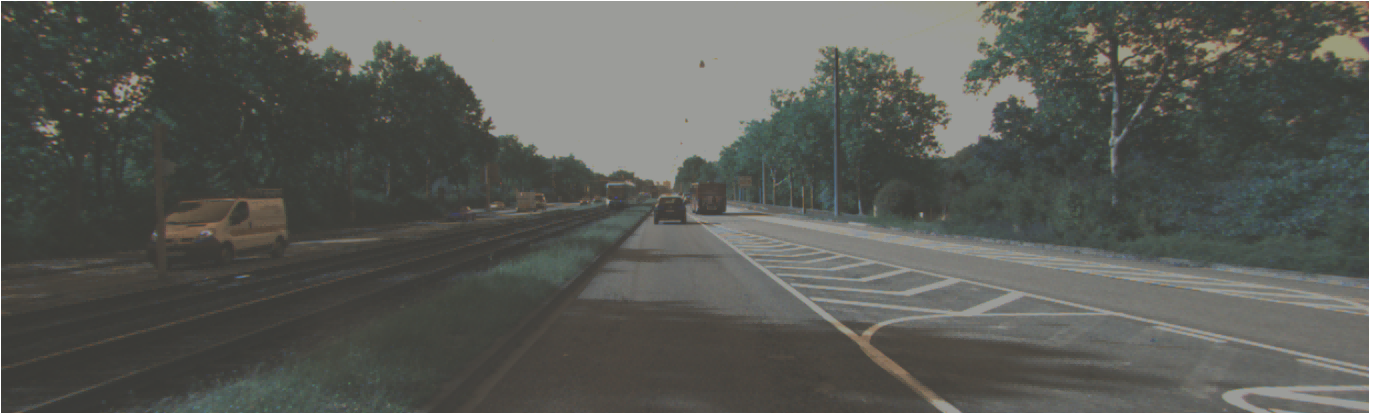}
        \caption*{Contrast}
        
    \end{subfigure}
    \hfill
    \begin{subfigure}[b]{0.3\textwidth}
        \centering
        \includegraphics[width=\linewidth]{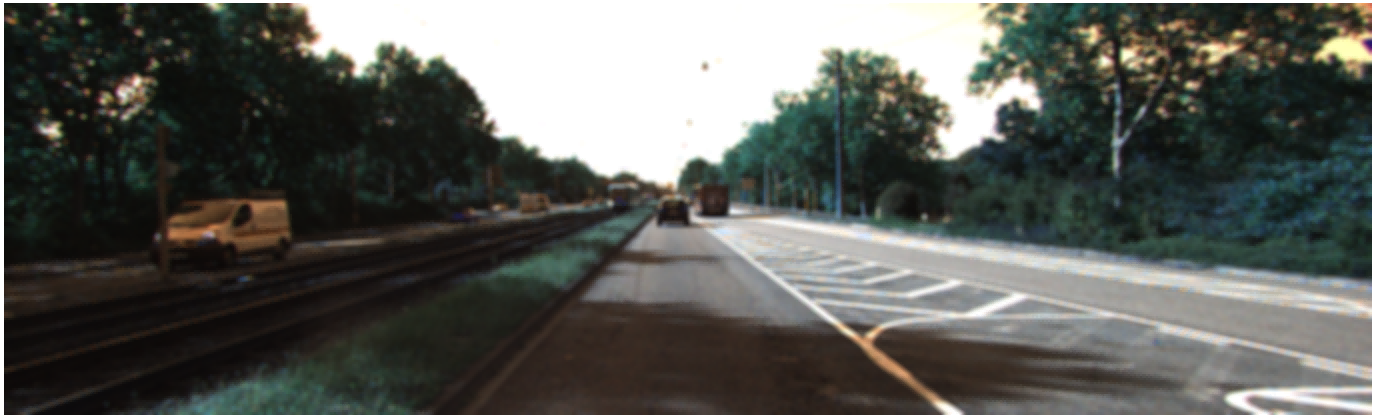}
        \caption*{Defocus}
        
    \end{subfigure}
    \hfill
    \begin{subfigure}[b]{0.3\textwidth}
        \centering
        \includegraphics[width=\linewidth]{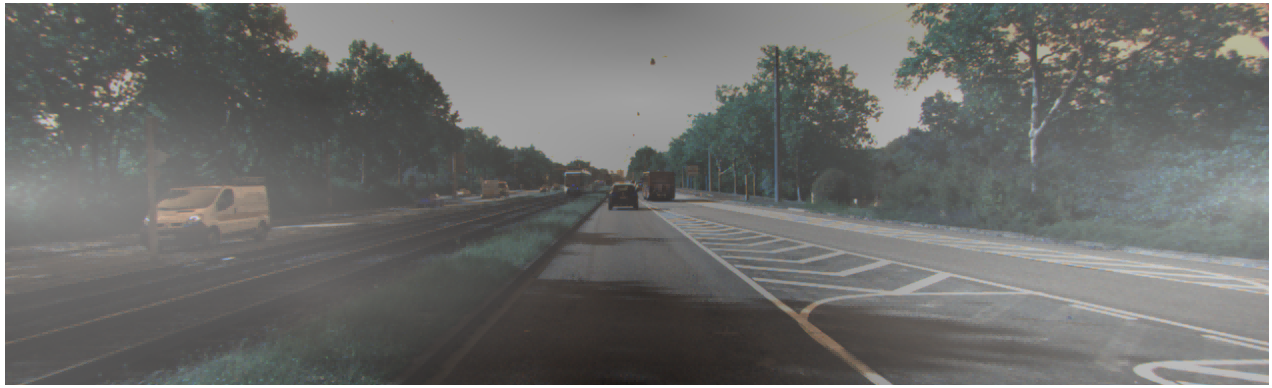}
        \caption*{Fog}
        
    \end{subfigure}
    \hfill
    \begin{subfigure}[b]{0.3\textwidth}
        \centering
        \includegraphics[width=\linewidth]{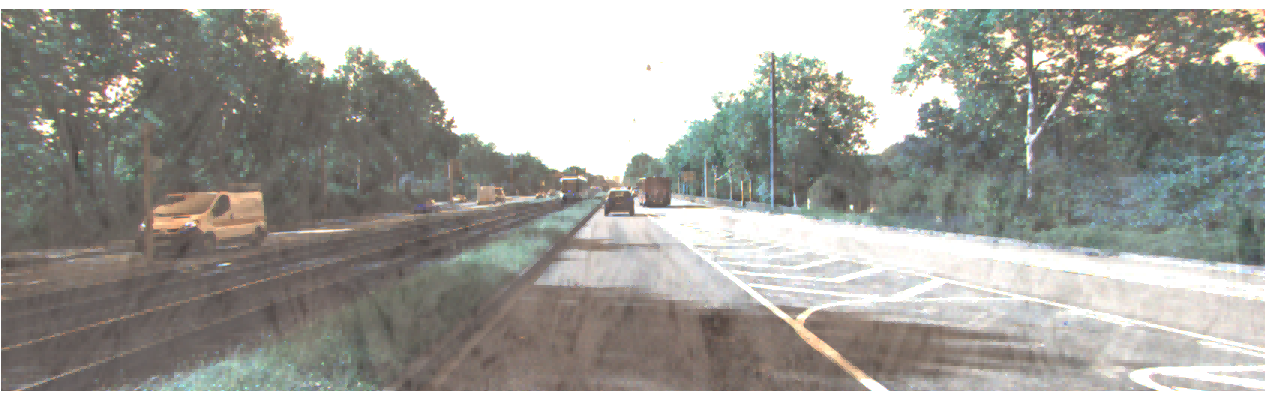}
        \caption*{Frost}
        
    \end{subfigure}
    \hfill
    \begin{subfigure}[b]{0.3\textwidth}
        \centering
        \includegraphics[width=\linewidth]{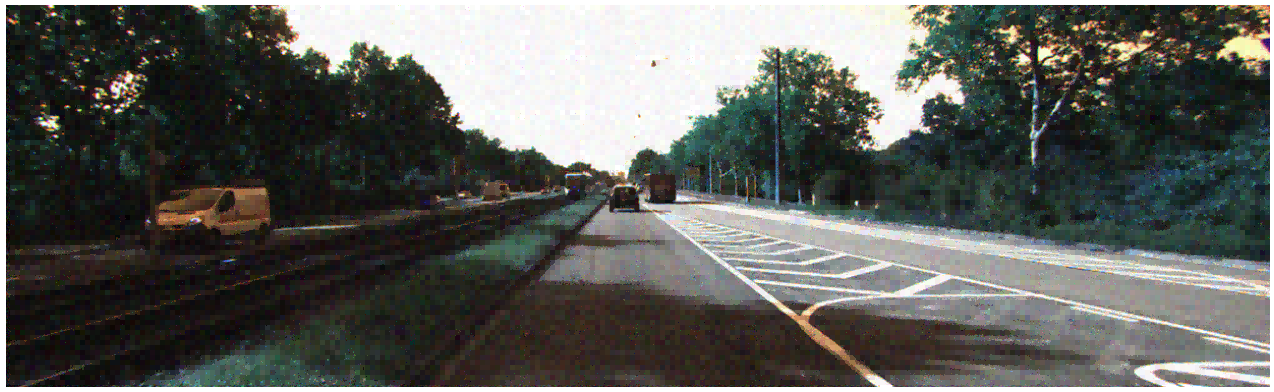}
        \caption*{Gaussian Noise}
        
    \end{subfigure}
    \hfill
    \begin{subfigure}[b]{0.3\textwidth}
        \centering
        \includegraphics[width=\linewidth]{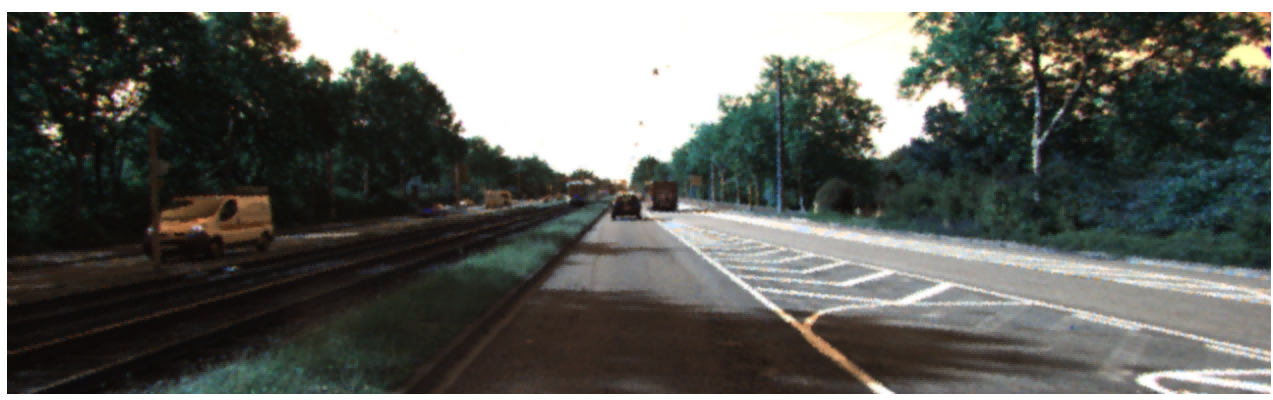}
        \caption*{Glass Blur}
        
    \end{subfigure}
    \hfill
    \begin{subfigure}[b]{0.3\textwidth}
        \centering
        \includegraphics[width=\linewidth]{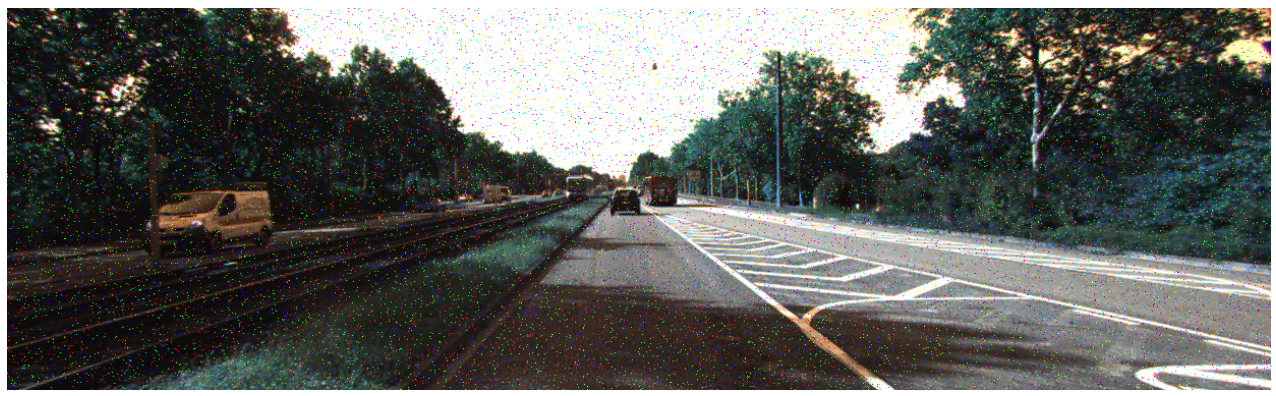}
        \caption*{Impulse Noise}
        
    \end{subfigure}
    \hfill
    \begin{subfigure}[b]{0.3\textwidth}
        \centering
        \includegraphics[width=\linewidth]{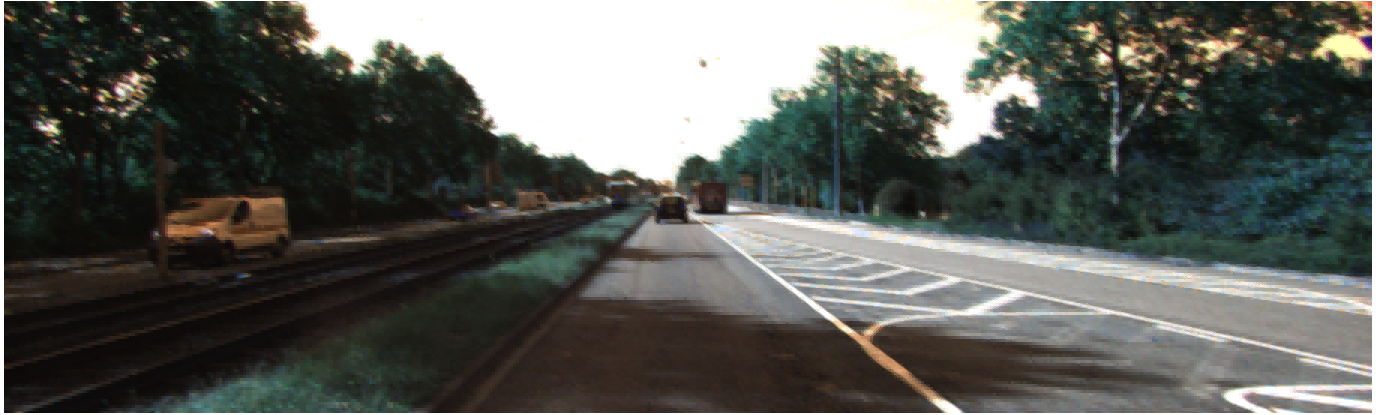}
        \caption*{Motion Blur}
        
    \end{subfigure}
    \hfill
    \begin{subfigure}[b]{0.3\textwidth}
        \centering
        \includegraphics[width=\linewidth]{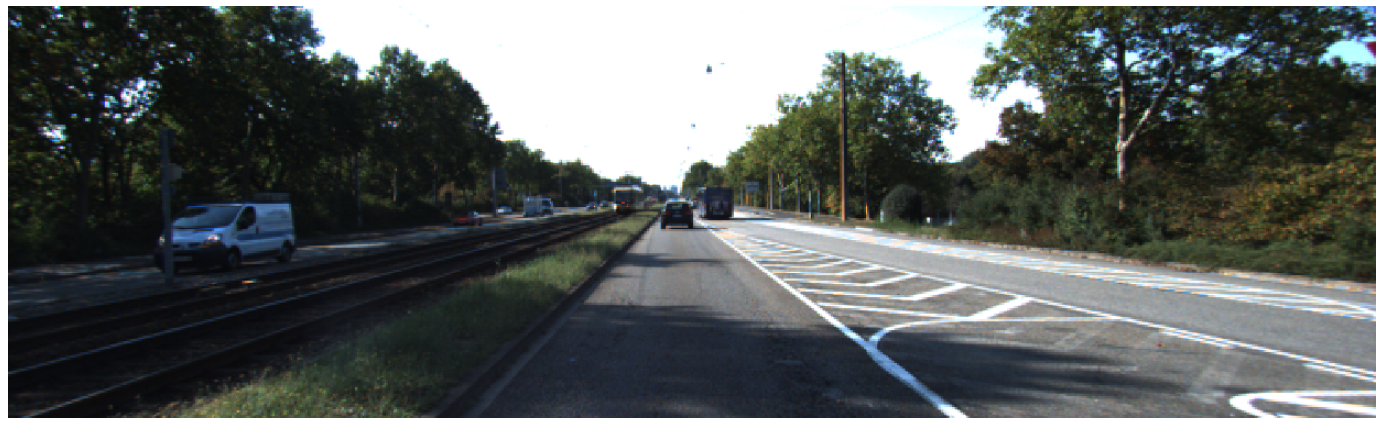}
        \caption*{Pixelate}
        
    \end{subfigure}
    \hfill
    \begin{subfigure}[b]{0.3\textwidth}
        \centering
        \includegraphics[width=\linewidth]{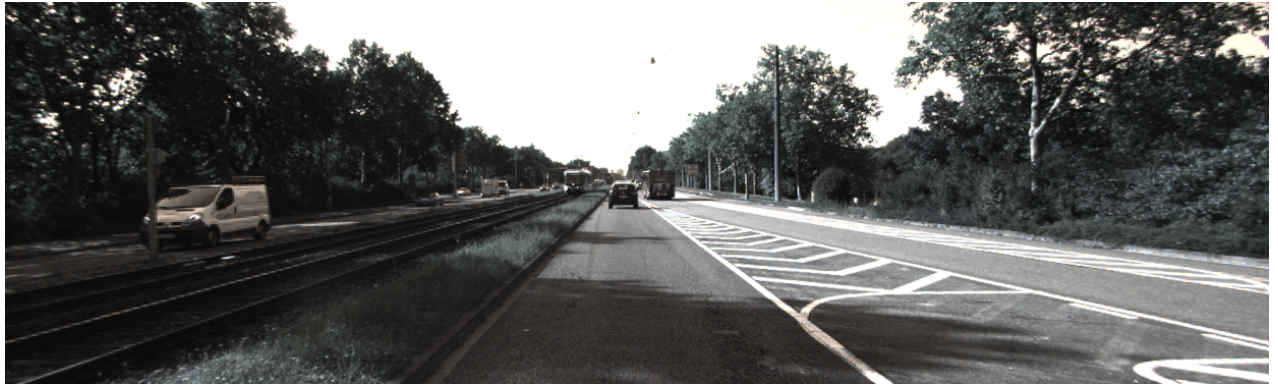}
        \caption*{Saturate}
       
    \end{subfigure}
    \hfill   
    \begin{subfigure}[b]{0.3\textwidth}
        \centering
        \includegraphics[width=\linewidth]{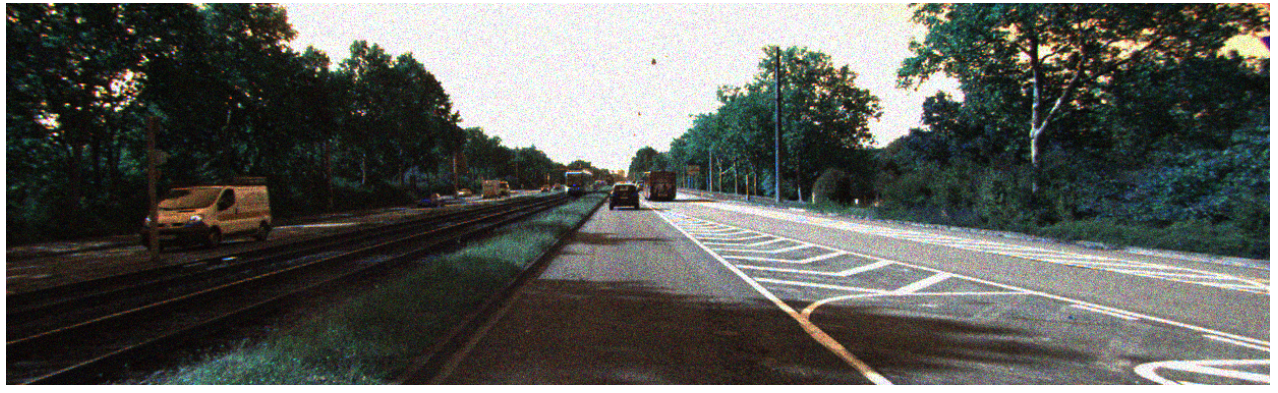}
        \caption*{Shot Noise}
        
    \end{subfigure}

    \caption{the 12 distinct types of corruptions in Kitti-C dataset}
    
    \label{Kitti-C quant}
    
\end{figure*}

As shown in Fig~\ref{Nuscenes-c vis}, We present visualization results for six different types of corruption across three  severity levels from the NuScenes-C dataset,  As shown in Fig~\ref{Kitti-C quant} is the visualization results for twelve types of corruption from the KITTI-C dataset. 

\begin{figure*}
    \centering
    \begin{subfigure}[b]{\textwidth}
        \centering
        \includegraphics[width=\linewidth]{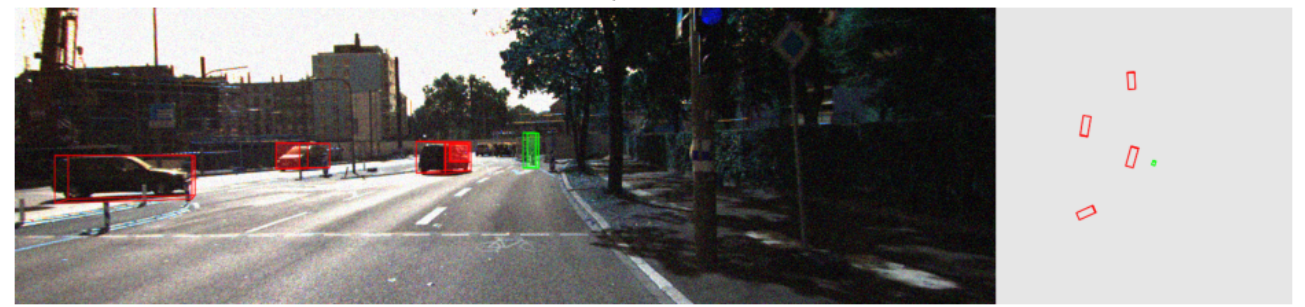}
        \caption{LearnableBN}
        
    \end{subfigure}
    \hfill
    \begin{subfigure}[b]{\textwidth}
        \centering
        \includegraphics[width=\linewidth]{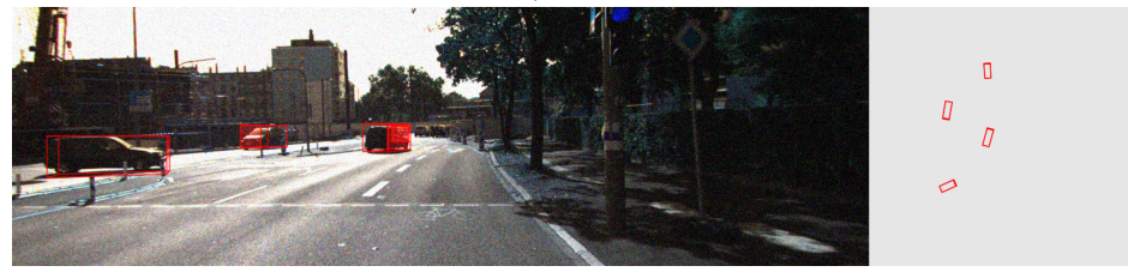}
        \caption{Baseline}
        
    \end{subfigure}
    \caption{Visualization examples from the \textbf{Gaussian Noise} scenario in the KITTI-C dataset, illustrating the differences before and after applying our proposed LearnableBN method. the red box is ground truth, the green box is detection results.}
    \label{Kitti-C vis}
\end{figure*}

As shown in Fig~\ref{Kitti-C vis}, we also demonstrate the results obtained using the LearnableBN method in the Gaussian Noise corruption scenario of the KITTI-C dataset. Consistent with the previous experimental results, LearnableBN significantly improves the baseline model’s prediction for the pedestrian category. This indicates that the LearnableBN method enhances the model's robustness across multiple categories and alleviates the long-tail effect.

\begin{figure*}
    \centering
    \begin{subfigure}[b]{\textwidth}
        \centering
        \includegraphics[width=\linewidth]{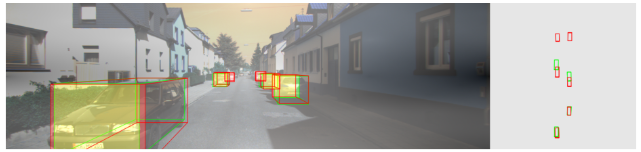}
        \caption{LearnableBN}
        
    \end{subfigure}
    \hfill
    \begin{subfigure}[b]{\textwidth}
        \centering
        \includegraphics[width=\linewidth]{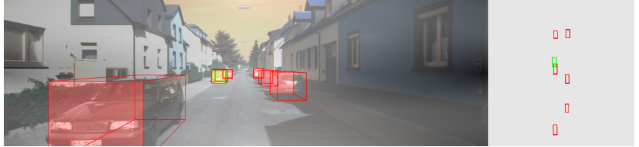}
        \caption{Baseline}
        
    \end{subfigure}
    \caption{Visualization examples from the \textbf{Fog} scenario in the KITTI-C dataset, illustrating the differences before and after applying our proposed LearnableBN method. the red box is ground truth, the green box is detection results.}
    \label{Kitti-C vis fog}
\end{figure*}

\begin{figure*}
    \centering
    \begin{subfigure}[b]{\textwidth}
        \centering
        \includegraphics[width=\linewidth]{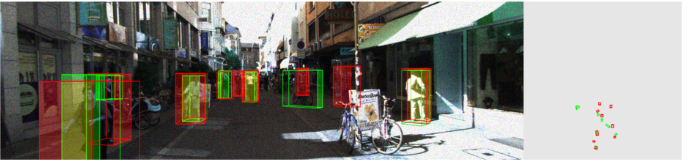}
        \caption{LearnableBN}
        
    \end{subfigure}
    \hfill
    \begin{subfigure}[b]{\textwidth}
        \centering
        \includegraphics[width=\linewidth]{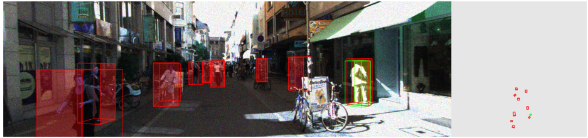}
        \caption{Baseline}
        
    \end{subfigure}
    \caption{Visualization examples from the \textbf{Shot Noise} scenario in the KITTI-C dataset, illustrating the differences before and after applying our proposed LearnableBN method. the red box is ground truth, the green box is detection results.}
    \label{Kitti-C vis shot}
\end{figure*}

\begin{figure*}
    \centering
    \begin{subfigure}[b]{\textwidth}
        \centering
        \includegraphics[width=\linewidth]{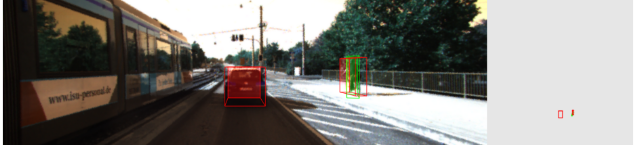}
        \caption{LearnableBN}
        
    \end{subfigure}
    \hfill
    \begin{subfigure}[b]{\textwidth}
        \centering
        \includegraphics[width=\linewidth]{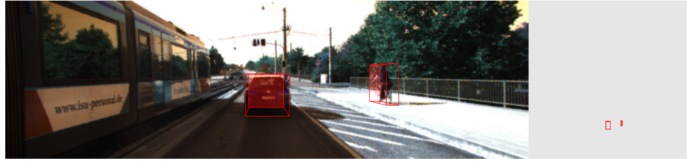}
        \caption{Baseline}
        
    \end{subfigure}
    \caption{Visualization examples from the \textbf{Glass Blur} scenario in the KITTI-C dataset, illustrating the differences before and after applying our proposed LearnableBN method. the red box is ground truth, the green box is detection results.}
    \label{Kitti-C vis glass}
\end{figure*}

\begin{figure*}
    \centering
    \begin{subfigure}[b]{\textwidth}
        \centering
        \includegraphics[width=\linewidth]{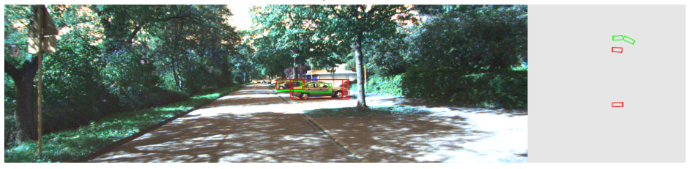}
        \caption{LearnableBN}
        
    \end{subfigure}
    \hfill
    \begin{subfigure}[b]{\textwidth}
        \centering
        \includegraphics[width=\linewidth]{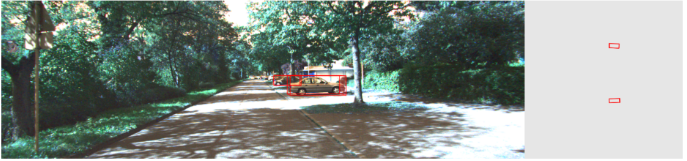}
        \caption{Baseline}
        
    \end{subfigure}
    \caption{Visualization examples from the \textbf{Brightness} scenario in the KITTI-C dataset, illustrating the differences before and after applying our proposed LearnableBN method. the red box is ground truth, the green box is detection results.}
    \label{Kitti-C vis bright}
\end{figure*}